\documentclass[letterpaper]{article} 
\usepackage{aaai2026}  
\usepackage{times}  
\usepackage{helvet}  
\usepackage{courier}  
\usepackage[hyphens]{url}  
\usepackage{graphicx} 
\usepackage{natbib}  
\usepackage{caption} 
\usepackage{algorithm}
\usepackage{algorithmic}
\usepackage{booktabs}
\usepackage{svg}
\usepackage{adjustbox}
\usepackage{tikz}
\usetikzlibrary{shapes, positioning, shadows}
\usepackage{multirow}
\usepackage{soul}
\usepackage{makecell}
\usepackage{tabularx}
\newcolumntype{Y}{>{\centering\arraybackslash}X}
\usepackage[T1]{fontenc}
\usepackage{newfloat}
\usepackage{listings}
\DeclareCaptionStyle{ruled}{labelfont=normalfont,labelsep=colon,strut=off} 
\floatstyle{ruled}
\newfloat{listing}{tb}{lst}{}
\floatname{listing}{Listing}

\title{RetrySQL: Text-to-SQL Training with Retry Data\\for Self-Correcting Query Generation}
\author{
    Alicja Rączkowska\equalcontrib, Riccardo Belluzzo\equalcontrib, Piotr Zieliński\equalcontrib, Joanna Baran\equalcontrib, Paweł Olszewski\equalcontrib
}
\affiliations{
    Machine Learning Research\\

    Allegro.com\\
    \{alicja.raczkowska, riccardo.belluzzo, piotr.c.zielinski, joanna.baran, pawel.olszewski\}@allegro.com
}

\begin{document}

\maketitle

\begin{abstract}
The text-to-SQL task is an active challenge in Natural Language Processing. Many existing solutions focus on using black-box language models extended with specialized components within customized end-to-end text-to-SQL pipelines. While these solutions use both closed-source proprietary language models and coding-oriented open-source models, there is a lack of research regarding SQL-specific small generative models. At the same time, recent advancements in self-correcting generation strategies show promise for improving the capabilities of existing architectures. The application of these concepts to the text-to-SQL task remains unexplored.
In this paper, we introduce \textit{RetrySQL}, a new approach to training text-to-SQL generation models. We prepare reasoning steps for reference SQL queries and then corrupt them to create \textit{retry data} that contains both incorrect and corrected steps, divided with a special token. We continuously pre-train open-source coding models with this data and demonstrate that retry steps yield an improvements of up to 4 and 9 percentage points for overall and challenging execution metrics, respectively, as compared to pre-training without \textit{retry data}. We showcase that the self-correcting behavior is learned by the model and the increase in downstream accuracy metrics is a result of this additional skill. Finally, we incorporate \textit{RetrySQL}-trained models into the full text-to-SQL pipeline and showcase that they are competitive in terms of Execution Accuracy with proprietary models that contain orders of magnitude more parameters.
\textit{RetrySQL} demonstrates that self-correction can be learned in the text-to-SQL task and provides a novel way of improving generation accuracy for small SQL-oriented language models.
\end{abstract}

\begin{links}
    \link{Code}{https://github.com/allegro/RetrySQL}
    \link{Extended version}{https://arxiv.org/abs/2507.02529}
\end{links}

\section{Introduction}

The task of translating natural language questions to SQL queries is a major challenge for machine learning models. The complexity stems from the need of relating often ambiguous user input to abstract entities, relations and values that are present in relational databases~\cite{li_can_2023}. Even prominent Large Language Models (LLMs), such as GPT-4o~\cite{openai_2024} or Gemini 1.5~\cite{gemini_2024}, struggle with approaching human performance in leading text-to-SQL benchmarks: BIRD~\cite{li_can_2023} and SPIDER 2.0~\cite{lei_spider2_2024}. The same is true for models tuned specifically for coding tasks \cite{li_codes_2024, chen2024_opensql, talaei_chess_2024}. Thus, there exists a need for more advanced solutions that can bridge that gap and provide reliable SQL queries even in difficult real-world scenarios. 

The text-to-SQL task can be divided into three main steps~\cite{maamari_death_2024}: retrieval, generation and correction. Many existing approaches try to tackle these steps at the same time, in a single end-to-end pipeline~\cite{maamari_death_2024,talaei_chess_2024,pourreza_din-sql_2023,pourreza_dts-sql_2024}, utilizing relatively large LLMs. In this work, we focus only on the generation step and show how it can be improved with a novel approach to model pre-training of small LLMs - text-to-SQL systems need to operate quickly in real-world scenarios.

\begin{figure*}[t]
  \centering
  \includegraphics[width=0.89\textwidth,trim={0.8cm 0 0 0},clip]{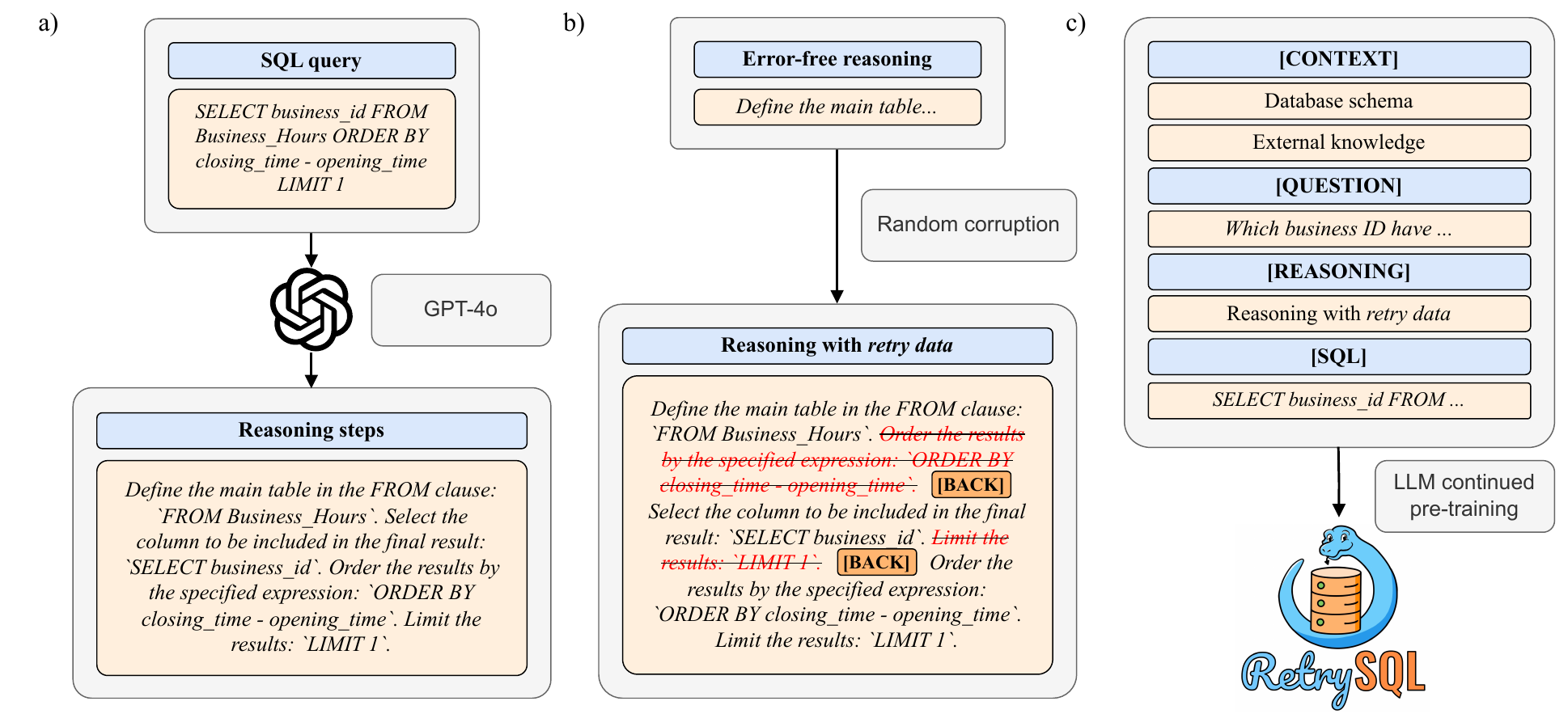}
  \caption{\textit{RetrySQL} overview. \textbf{(a)} Reasoning step generation. For each SQL query in the training dataset, we generate a series of reasoning steps using GPT-4o. \textbf{(b)} Preparation of \textit{retry data}. For each set of reasoning steps, we apply random perturbations, treated as errors, by replacing some steps with different ones. We follow these errors with special \textbf{[BACK]} tokens and amend them with correct steps. \textbf{(c)} We take an open-source LLM and continue its pre-training with training examples that contain \textit{retry data} injected into reasoning steps. The~resulting \textit{RetrySQL}-trained model learns the ability to self-correct, which improves its capabilities in generating correct SQL queries from natural language questions.}
  \label{fig:teaser}
\end{figure*}

Specifically, we teach the model to self-correct during the generation itself. While previous work did use self-correction in the sense of post-processing, applied in the correction step~\cite{pourreza_din-sql_2023}, we enforce the self-correcting behavior at an earlier point. This sort of active knowledge-based self-correction is an ongoing research area when it comes to LLMs~\cite{ye_physics_part2point1_2024, ye_physics_part2point2_2024}. While recent work in slow thinking reasoning systems, such as DeepSeek-R1~\cite{deepseekai_2025}, shows that self-correction can be learned in a reinforcement learning setup, other lines of research suggest that it is possible to obtain the self-correction ability with specific data augmentations and a standard auto-regressive pre-training objective~\cite{ye_physics_part2point1_2024, ye_physics_part2point2_2024} or during supervised fine-tuning~\cite{muennighoff_s1_2025}. It has been shown that augmenting training data for grade-school math solution generation with so-called \textit{retry data} leads to increased generation accuracy~\cite{ye_physics_part2point2_2024}. The applicability of this approach to other tasks and models has not been explored as of yet.

We introduce \textit{RetrySQL}, a novel text-to-SQL generation module training paradigm that incorporates \textit{retry data} in the training process and teaches the resulting model to self-correct. \textit{RetrySQL} first augments the training data with reasoning steps that explain the sequence of operations required for obtaining the solution SQL query~(\textbf{Fig.~\ref{fig:teaser}a}). Then, \textit{retry data} is generated by corrupting the order of these reasoning steps~(\textbf{Fig.~\ref{fig:teaser}b}). The \textit{retry data} is incorporated into the training examples and we perform continued pre-training of open-source coding-oriented LLMs, which results in a \textit{RetrySQL}-trained models that are capable of self-correction~(\textbf{Fig.~\ref{fig:teaser}c}). We present a set of experiments across multiple strategies of generating reasoning corruptions. We demonstrate that using \textit{retry data} yields superior generation results when compared to training data with error-free reasoning steps.

To showcase that the self-correcting behavior is indeed learned by the model, we provide an analysis of output token confidence. 
We show that the max softmax score is on average lower for tokens before correction than for those after. Similarly, incorrect tokens display higher variance of softmax scores across beam search passes than the corrected ones. This illustrates that the model becomes uncertain as it makes a mistake and then self-corrects itself with higher confidence.


Our results corroborate the recent findings regarding the self-correction ability in language models~\cite{ye_physics_part2point2_2024}, demonstrating that the improvement in generation accuracy coming from the inclusion of \textit{retry data} in pre-training is a universal law. It is applicable not only to the grade school math reasoning problem and GPT-2, but also to the text-to-SQL domain and larger, more modern Transformer-based decoder-only models. These findings suggest that \textit{retry data} could be adapted to even more domains, especially if reasoning steps can be added to the training examples.

While the main focus of our work is on the SQL generation step in isolation, we also showcase that relatively small, 1.5B-parameter open-source 
coding models trained with our \textit{RetrySQL} paradigm are competitive with much larger closed-source proprietary LLMs when used as a part of the full text-to-SQL pipeline.
We share all code for \textit{retry data} generation, as well as model training and evaluation.

In summary, our key contributions are the following:
\begin{itemize}
    \item We introduce \textit{RetrySQL}, a novel text-to-SQL training paradigm that makes use of reasoning steps enhanced with \textit{retry data}.
    \item We show that using \textit{retry data} in pre-training is beneficial to the generation process, as indicated by the Execution Accuracy metric calculated for the BIRD and SPIDER benchmark datasets.
    \item We demonstrate that \textit{RetrySQL}-trained models have the ability to self-correct as they generate reasoning steps for the output SQL queries.
    \item We illustrate that within a simple end-to-end text-to-SQL pipeline, \textit{RetrySQL}-trained 1.5B-parameter open-source coding models are competitive with proprietary models such as GPT-4o-mini and GPT-4o, which opens new possibilities for employing such small models in real-world text-to-SQL systems.
\end{itemize}

\section{Related work}

Early text-to-SQL methods relied on sequence-to-sequence frameworks, using models like Graph Neural Networks, Recurrent Neural Networks, and pre-trained Transformers for encoding queries and schemas~\cite{cai_sadga_2021, cao_lgesql_2021, wikisql_2019}, while employing slot-filling or auto-regressive decoding to generate SQL queries~\cite{ryansql_2020, wang_rat-sql_2020}. Recently, the field has shifted with the emergence of LLMs, which are currently leading in the most popular benchmarks~\cite{shi_survey_2024}. 
While initial efforts were focused on optimizing prompt designs that leveraged in-context learning~\cite{nan2023enhancingfewshottexttosqlcapabilities, gao2023_dailsql} and multi-stage prompting~\cite{pourreza_din-sql_2023}, the current state-of-the-art is represented by LLM-based pipelines. These latest approaches integrate LLMs in more complex sequences of processing stages, with separate components for schema linking, self-correction, self-debugging, and self-consistency~\cite{talaei_chess_2024, pourreza2024_chasesql, lee2024_mscsql, maamari_death_2024, sun2024_sqlpalm}.

Compared to closed-source model prompting approaches, open-source model fine-tuning for the text-to-SQL task remains relatively unexplored~\cite{shi_survey_2024}. Many of the existing works favor parameter-efficient fine-tuning over full-parameter fine-tuning due to the former's superior training efficiency and lower training costs~\cite{shi_survey_2024, chen2024_opensql, zhang2024_sqlfuse, zhang2024_finsql}. While the majority of practitioners choose to use powerful general-purpose LLMs as their base models~\cite{pourreza_dts-sql_2024, xie2024_deasql, shi_survey_2024},
promising results have also been shown by adapting coding LLMs to the text-to-SQL domain~\cite{li_codes_2024, chen2024_opensql, talaei_chess_2024}, demonstrating that starting from a model already heavily pre-trained on coding tasks, with SQL-related training data, leads to higher performance in benchmark evaluations.

The usage of small LLMs for the text-to-SQL task is an important practical concern, as response time and costs are crucial factors in real-world applications. Existing work explored models as small as 7B parameters~\cite{pourreza_dts-sql_2024}, but the utilization of even smaller LLMs remains relatively unexplored.

It has been shown that language models trained to follow chain of thought (CoT)~\cite{wei_chain_2023} steps excel at solving problems that involve math and symbolic reasoning~\cite{GSM, sprague2024_tocot_or_nottocot, ye_physics_part2point1_2024, ye_physics_part2point2_2024, muennighoff_s1_2025}. Recent work in the domain of LLM theory paves the way for the discovery of LLM universal laws, 
including the ability to self-correct~\cite{ye_physics_part2point1_2024, ye_physics_part2point2_2024}. Test-time compute methods are an active area of research as well, with some methods indicating that reinforcement learning approaches might not be needed for eliciting self-reflection in LLMs~\cite{madaan_self-refine_2023, muennighoff_s1_2025}. When it comes to small reasoning models, it has been shown that data curation~\cite{gunasekar_textbooks_2023} and knowledge distillation~\cite{mitra_orca2_2023} are effective techniques for improving model performance.

The intersection of the text-to-SQL task and reasoning with self-correction for small ($\sim$1B parameters) LLMs has not been explored as of yet. \textit{RetrySQL} addresses this gap by providing a method for teaching such small text-to-SQL models to self-correct, which greatly improves their capabilities.


\section{Methodology}

In this section, we describe our \textit{RetrySQL} training paradigm in detail. We augment the training dataset (\textbf{Section~\ref{paragraph: training_data}}) with synthetically generated reasoning steps (\textbf{Section~\ref{paragraph: reasoning_data}}). Then, we define the \textit{retry data} generation process, in which reasoning steps are corrupted with random errors and then corrected (\textbf{Section~\ref{paragraph: retry_data}}). 

\subsection{Training data}
\label{paragraph: training_data}

In order to assess the efficacy of RetrySQL, we utilized two popular text-to-SQL benchmark datasets: BIRD~\cite{li_can_2023} and SPIDER~\cite{yu_spider_2019}. We selected them due to their relatively large training sets. For BIRD, the training data includes 9428 examples, each consisting of the database name, natural language question, external evidence, and the ground truth SQL query (SQLite dialect). In addition, the metadata for each database is available as well, consisting of a full list of tables, columns and table relations. We discovered that relations for one table, \textit{mondial\_geo}, are defined incorrectly. We excluded it from our pipeline, which left us with 9135 training examples. For SPIDER, the training data consists of 8659 examples, with a similar structure to BIRD, excluding the external evidence. We did not use the newer SPIDER 2.0~\cite{lei_spider2_2024} dataset due to its focus on agentic evaluation - it does not include any training examples.

For the generation process, we needed to incorporate the schema information together with the question and external knowledge. To this end, we parsed the ground truth queries and prepared the matching Data Definition Language (DDL) statements, which served as the schema linking data. Importantly, unless stated otherwise, we incorporated so-called \textit{perfect} schema linking in our experiments (i.e. no redundant links). We were interested primarily in studying the SQL generation process in isolation, without the additional task of finding relevant schema connections. We matched column and table names in each ground truth SQL query to the corresponding database metadata and built minimal required schema links. 

We used DDL for schema representation because it provides a concise notation that includes table and column names, together with data types and relations. Moreover, it keeps an SQL-focused context for the model, without needing to explain specific data formats in the prompt. This approach is commonly used in existing text-to-SQL pipelines \cite{shi_survey_2024}.

\subsection{Reasoning step generation}
\label{paragraph: reasoning_data}

Previous work showed that the usage of \textit{retry data} in model pre-training is effective only if we also instruct the model to follow a chain of reasoning steps~\cite{ye_physics_part2point2_2024}. To this end, we needed to procure reasoning steps for each of our training examples. The training datasets do not contain this data, so we used GPT-4o for generating synthetic reasoning steps (\textbf{Fig.~\ref{fig:teaser}a}). Enhancing language model training data with synthetically generated components is a newly emerging trend~\cite{abdin_phi_2024}. We used a prompt that highlighted the need of reasoning steps being in a format resembling solution reasoning chains from the dataset used in previous research on self-correction~\cite{ye_physics_part2point1_2024}~(\textbf{Fig.~S3}). We verified the correctness of the output formatting and also the semantic validity for a small subset of all examples, consisting of 100 instances sampled uniformly at random, thus representing a wide range of databases and query difficulties. We found that there were no cases with erroneous reasoning steps. Consequently, we assumed that the full dataset is similarly error-free. Full manual verification was not necessary, since we did not aim for the reasoning steps to be perfectly accurate. We ultimately wanted to generate SQL queries, and the reasoning steps were meant to serve as an additional training signal to the verified ground truth from the training datasets.

\subsection{Retry data generation}
\label{paragraph: retry_data}

We generated SQL-specific \textit{retry data} by corrupting the solution steps prepared beforehand~\cite{ye_physics_part2point2_2024} (\textbf{Fig.~\ref{fig:teaser}b}). We considered several variants of perturbations: forward single (denoted as \textit{FS}), forward and back single (\textit{FBS}), forward multiple (\textit{FM}), forward and back multiple (\textit{FBM}). Given a sequence of reasoning steps of length~$N$, for each step $r_i$ in that sequence we select uniformly at random (with probability $p_{retry}$) another step $r_{error}\in S$, where $S$~is a set of candidate corruptions. The selection is done either once (for \textit{FS} and \textit{FBS}) or multiple times (for \textit{FM} and \textit{FBM}). For \textit{FS} and \textit{FM}, $S$ consists of elements $r_{i+1},...\ ,r_N$ (i.e. future steps). For \textit{FBS} and \textit{FBM}, $S$~contains elements $r_1,...\ ,r_{i-1},r_{i+1},...\ ,r_N$ (i.e. future and past steps). After selecting an element $r_{error}$ from $S$, we follow it with the \textbf{[BACK]} token and then with the correct $r_i$ itself. As such, each $r_i$ can be replaced with the following:
\begin{itemize}
    \item ($r_{error}$, \textbf{[BACK]}, $r_i$), for \textit{FS} and \textit{FBS},
    \item ($r_{error}$, \textbf{[BACK]}, ... , $r_{error}$, \textbf{[BACK]}, $r_i$), for \textit{FM} and \textit{FBM}.
\end{itemize}

We generated training dataset variants for \textit{FS}, \textit{FBS}, \textit{FM}, \textit{FBM} and $p_{retry}\in \{0.1, 0.2, 0.3, 0.4, 0.5\}$. To steer the training, we introduced additional tokens \textbf{[CONTEXT]}, \textbf{[QUESTION]}, \textbf{[REASONING]} and \textbf{[SQL]}. We combined the database schema, external knowledge, reasoning steps and the ground truth SQL query using these tokens as delimiters~(\textbf{Fig.~S2}).


\section{Experiments}

In this section, we outline the experimental setup for our study. We delineate the preliminary linear probing task (\textbf{Section~\ref{paragraph: linear_probing_data}}), and then describe the baseline models that we used in our experiments (\textbf{Section~\ref{paragraph: baseline_models}}). We explain our inference procedure, as well as evaluation metrics used to measure the effectiveness of all models (\textbf{Section~\ref{paragraph: inference}}).

\subsection{Baseline models}
\label{paragraph: baseline_models}

We evaluated several baselines in addition to the models trained with \textit{retry data}. We assessed zero-shot performance of GPT-4o-mini\footnote[1]{API version: 2023-03-15-preview}, GPT-4o\footnotemark[1], Gemini-1.5-flash\footnote[2]{Stable version: 002} and Gemini-1.5-pro\footnotemark[2], as well as several open-source models: Llama-3.2~1.5B, Qwen2.5~1.5B, Qwen2.5-Coder~1.5B~\cite{qwen_2.5_2025} and OpenCoder~1.5B~\cite{huang_opencoder_2024}.

For our experiments with \textit{retry data}, we chose from among them two coding-oriented models: OpenCoder~1.5B and Qwen2.5-Coder~1.5B. Such small model sizes allowed us to more effectively utilize our compute resources and sped up the experimentation process. 


For the details on the training setup and hyperparameters, see \textbf{Appendix~A.1}. For the specifics of proprietary model inference, see \textbf{Appendix~A.2}.


\subsection{Linear probing dataset}
\label{paragraph: linear_probing_data}

Before conducting the pre-training experiments with \textit{retry data}, we first validated if the OpenCoder model has an innate, hidden capability of distinguishing \textit{correct} and \textit{incorrect} reasoning steps. If that were the case, then providing \textit{retry data} at training time would allow the model to unlock the ability to self-correct. It has been shown previously that models pre-trained on grade school math data with correct solution steps exhibit regretful patterns in their internal states \cite{ye_physics_part2point2_2024}.

In order to verify the above hypothesis, we designed a preliminary linear probing task. We parsed the \textit{retry data} and categorized the training examples based on the presence of the \textbf{[BACK]} token. Each reasoning step $r{_i}$ was marked as either:
\begin{itemize}
    \item \textit{incorrect}, if it was followed by the \textbf{[BACK]} token;
    \item \textit{correct}, if it was followed by another step.
\end{itemize} 

Then, we took the original BIRD data samples and extended them with reasoning step sequences ending with either \textit{correct} or \textit{incorrect} steps, again with the addition of special \textbf{[CONTEXT]}, \textbf{[QUESTION]} and \textbf{[REASONING]} tokens to divide the input sections. We used the Retry \textit{FS} 0.3 dataset variant as the source for the reasoning steps. In this way, we extracted $15k$ examples in total, keeping the proportion between \textit{correct} and \textit{incorrect} instances balanced. 

We used this data to train a classification model, in which a binary classification head categorized \textit{correct} and \textit{incorrect} reasoning steps (for more details, see \textbf{Appendix~A.6}).

\subsection{Inference process and evaluation metrics}
\label{paragraph: inference}

For the purpose of measuring the effectiveness of \textit{retry data} in text-to-SQL generation, we utilized the Execution Accuracy (EX) metric. For BIRD, the development dataset contains a total of 1534 examples. The data format is the same as the training set, with the addition of a difficulty value for each example. For SPIDER, the development data numbers 1034 examples, each with a difficulty level as well. 

During inference, we used the same format as during training, but omitted the part of each sequence after the \textbf{[REASONING]} token. Thus, we wanted the model to first generate the reasoning steps, and then the \textbf{[SQL]} token, followed by the actual SQL query. 

During inference we used the best checkpoint of each trained model variant. We used beam search multinomial sampling with 4 beams as a decoding strategy (following~\cite{ye_physics_part2point2_2024}), with $temperature=0.5$, $top\_k=50$ and $top\_p=1.0$. We limited the number of new tokens to 1024. Each evaluation example was processed 5 times, for the purpose of measuring the model variance. The results were post-processed by removing the \textbf{[SQL]} and preceding tokens, leaving only the SQL query.

We then calculated the EX metric in the following manner. First, the generated SQL query as well as the ground truth query for a given question were executed in an SQLite database containing the development data. Then, resulting sets of rows were compared and the ratio of matching rows was saved. Finally, the match ratios for all examples were averaged. For BIRD, we report EX values over all examples, or over just the simple, moderate or challenging ones. For SPIDER, we report five EX values: overall, easy, medium, hard and extra hard.

\section{Results}

In this section, we assess the effectiveness of \textit{RetrySQL} for training SQL generation models. We showcase the baseline results for both proprietary and open-source LLMs (\textbf{Section~\ref{paragraph: res_baselines}}). We demonstrate through linear probing that the baseline OpenCoder model can recognize \textit{incorrect} reasoning steps (\textbf{Section~\ref{paragraph: res_linear_probing}}). We then show that training with \textit{retry data} improves Execution Accuracy compared to training with error-free reasoning steps (\textbf{Section~\ref{paragraph: res_pretrain_retry}}). 
We also explain the self-correction behavior with an analysis of model confidence around the \textbf{[BACK]} tokens (\textbf{Section~\ref{paragraph: res_confidence}}). 
Finally, we describe the evaluation of the full text-to-SQL pipeline (\textbf{Section~\ref{paragraph: res_pipeline}}).

\subsection{Zero-shot baselines}
\label{paragraph: res_baselines}

We evaluated a selection of models in zero-shot mode (\textbf{Tab.~S1}). We observe that all small open-source models fall behind proprietary models, with Qwen2.5-Coder 1.5B being the closest to GPT-4o-mini. General-purpose models (Llama-3.2 1B and Qwen2.5 1.5B) scored worse than coding-specific models (Qwen2.5-Coder 1.5B and OpenCoder 1.5B). 
Based on these results we selected Qwen2.5-Coder 1.5B and OpenCoder 1.5B for further experiments.

\begin{table*}[t]
\caption{Execution Accuracy for OpenCoder 1.5 and Qwen2.5-Coder 1.5 trained with \textit{RetrySQL}. All results are expressed in percentages, with mean and standard deviation over 5 multinomial beam search generations. The best results are marked in bold. Results with \textit{retry data} that improve upon the error-free training are indicated with an underline.}
\label{tab:results_pretraining_best}
\setlength\extrarowheight{0pt}
\centering
\aboverulesep = 0.5pt
\belowrulesep = 1.0pt
\begin{tabularx}{\textwidth}{|c|*{6}{Y}}
    \toprule
    \multirow{8}{*}[0em]{\rotatebox[origin=c]{90}{OpenCoder 1.5B}} & \multicolumn{6}{c}{
        \begin{tabularx}{478pt}{Y|*{3}{Y}|Y} 
        \multicolumn{5}{c}{BIRD} \\
        \midrule
        Dataset variant & EX$_{simple}$ & EX$_{moderate}$ & EX$_{challenging}$ & EX$_{overall}$ \\ 
        \midrule 
        -- (zero-shot) & 47.14 ± 0.45 & 27.63 ± 0.44 & 17.52 ± 0.55 & 38.44 ± 0.43 \\
        \midrule 
        error-free & 62.70 ± 0.07 & 43.53 ± 0.14 & 39.45 ± 0.28 & 54.71 ± 0.08 \\
        \midrule 
        Retry \textit{FS} 0.2 & \underline{\textbf{68.22 ± 0.12}} & \underline{\textbf{45.47 ± 0.14}} & \underline{40.28 ± 0.34} & \underline{\textbf{58.70 ± 0.09}} \\ 
        Retry \textit{FS} 0.3 & \underline{68.00 ± 0.00} & \underline{44.91 ± 0.26} & \underline{\textbf{43.31 ± 0.28}} & \underline{58.68 ± 0.06} 
        \end{tabularx}
    } \\
    & \multicolumn{6}{c}{
        \begin{tabularx}{478pt}{Y|*{4}{Y}|Y} 
        \midrule
        \multicolumn{6}{c}{SPIDER} \\
        \midrule
        Dataset variant & EX$_{easy}$ & EX$_{medium}$ & EX$_{hard}$ & EX$_{extra}$ & EX$_{overall}$ \\
        \midrule
        error-free & 90.24 ± 0.18 & 73.59 ± 0.10 & \textbf{69.88 ± 0.31} & 56.63 ± 0.00 & 74.25 ± 0.08 \\
        \midrule
        Retry \textit{FS} 0.2 & \underline{\textbf{91.53 ± 0.00}} & \underline{80.27 ± 0.00} & 66.09 ± 0.00 & \underline{\textbf{60.12 ± 0.27}} & \underline{\textbf{77.35 ± 0.04}} \\
        Retry \textit{FS} 0.3 & 89.60 ± 0.18 & \underline{\textbf{80.94 ± 0.00}} & 62.76 ± 0.26 & \underline{59.52 ± 0.27} & \underline{76.52 ± 0.08} \\
        \end{tabularx}
    } \\
    \midrule
    \multirow{4}{*}[0.5em]{\rotatebox[origin=c]{90}{Qwen2.5-Coder 1.5B}} & \multicolumn{6}{c}{
        \begin{tabularx}{478pt}{Y|*{3}{Y}|Y} 
        \multicolumn{5}{c}{BIRD} \\
        \midrule
        Dataset variant & EX$_{simple}$ & EX$_{moderate}$ & EX$_{challenging}$ & EX$_{overall}$ \\ 
        \midrule 
        -- (zero-shot) & 46.77 ± 0.44 & 24.78 ± 0.39 & 11.45 ± 0.94 & 36.78 ± 0.44 \\
        \midrule
        error-free & \textbf{65.47 ± 0.13} & 45.04 ± 0.00 & 30.21 ± 0.28 & 55.96 ± 0.10 \\
        \midrule
        Retry \textit{FS} 0.2 & 63.96 ± 0.09 & \underline{45.47 ± 0.00} & \underline{35.17 ± 0.00} & 55.65 ± 0.05 \\
        Retry \textit{FS} 0.3 & 63.91 ± 0.08 & \underline{\textbf{46.64 ± 0.22}} & \underline{\textbf{39.10 ± 0.00}} & \underline{\textbf{56.36 ± 0.05}}
        \end{tabularx}
    } \\
   & \multicolumn{6}{c}{
        \begin{tabularx}{478pt}{Y|*{4}{Y}|Y} 
        \midrule
        \multicolumn{6}{c}{SPIDER} \\
        \midrule
        Dataset variant & EX$_{easy}$ & EX$_{medium}$ & EX$_{hard}$ & EX$_{extra}$ & EX$_{overall}$ \\
        \midrule
        error-free & \textbf{92.34 ± 0.00} & 74.75 ± 0.11 & 65.52 ± 0.00 & 48.80 ± 0.00 & 73.25 ± 0.05 \\
        \midrule
        Retry \textit{FS} 0.2 & 90.40 ± 0.16 & \underline{\textbf{80.94 ± 0.00}} & \underline{\textbf{69.66 ± 0.43}} & \underline{\textbf{55.18 ± 0.30}} & \underline{\textbf{77.18 ± 0.06}} \\
        Retry \textit{FS} 0.3 & 91.13 ± 0.00 & \underline{77.76 ± 0.09} & \underline{66.55 ± 0.23} & \underline{53.01 ± 0.00} & \underline{75.11 ± 0.08} \\
        \end{tabularx}
    } \\
    \bottomrule
    
\end{tabularx}
\end{table*}

\subsection{Detecting regretful patterns through linear probing}
\label{paragraph: res_linear_probing}

\begin{figure}
    \centering
    \includegraphics[scale=0.32]{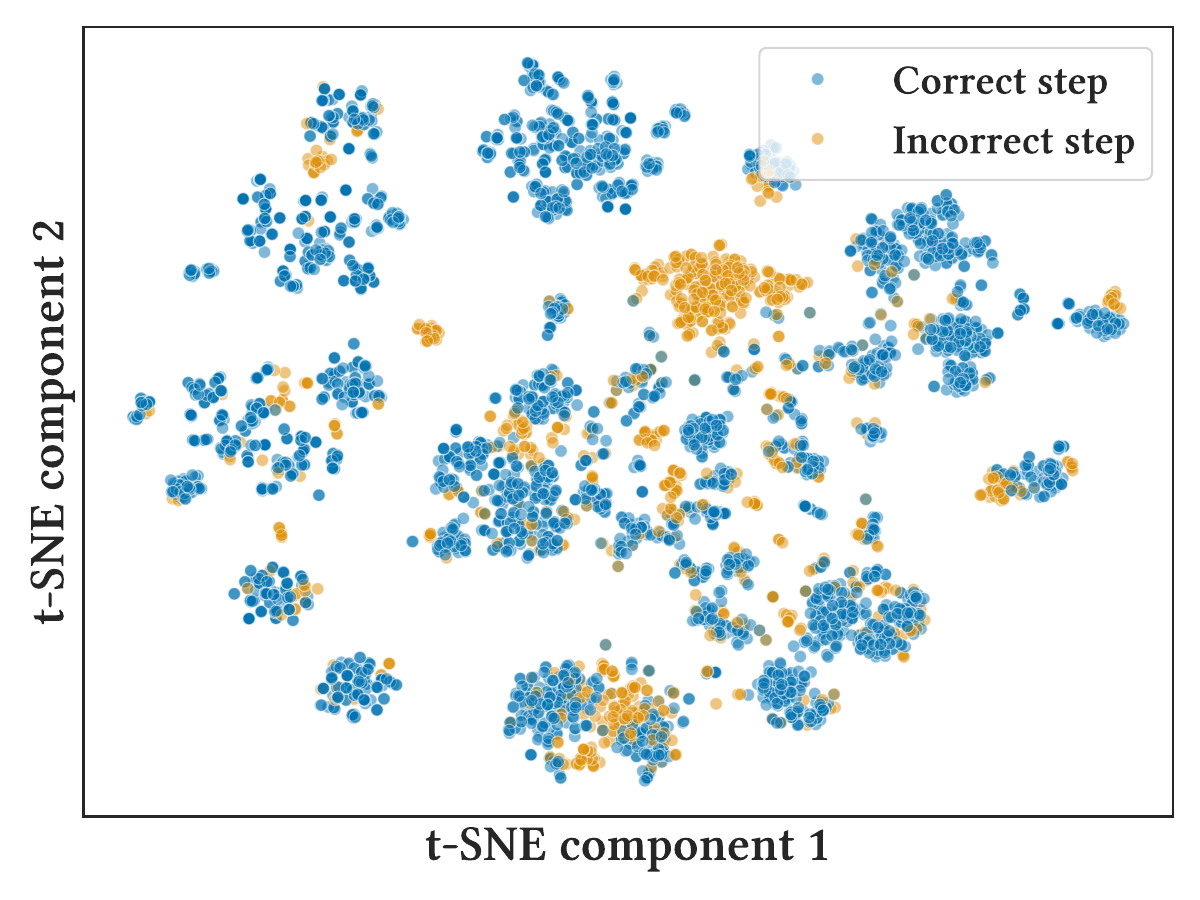}
    \caption{t-SNE projection of OpenCoder's internal state embeddings for the linear probing task. Blue points represent embeddings corresponding to \textit{correct} reasoning steps, while orange points indicate embeddings for \textit{incorrect} steps. The clusters of orange points indicate that the OpenCoder model differentiates a large portion of the \textit{incorrect} steps from the \textit{correct} ones, highlighting the innate, yet hidden, ability to detect mistakes in the reasoning process.}
    \label{fig:embeddings_plot}
\end{figure}

We took the baseline 
OpenCoder 1.5B model and performed a linear probing experiment with its frozen weights. The linear probing model detected \textit{incorrect} steps with average $balanced\_accuracy=82\%$ and $f1\_score=71\%$. Since the detection accuracy was significantly higher than $50\%$ (random guess), we can conclude that the probing signals most likely came from the pre-trained weights, and not from the fine-tuned classification layer~\cite{ye_physics_part2point1_2024}. To support our findings, we visualize the internal state embeddings of OpenCoder using a t-SNE projection (\textbf{Fig. \ref{fig:embeddings_plot}}). The plot illustrates the separability of the model's internal states when it comes to predicting \textit{correct} versus \textit{incorrect} reasoning steps. These results demonstrate that OpenCoder has an innate capability to self-correct, which is in line with previous research positing that this ability is a universal law for all Transformer models~\cite{ye_physics_part2point2_2024}. This justifies the usage of \textit{retry data}, which should enable the model to backtrack as it generates an~\textit{incorrect} step and then \textit{retry} once again.

\subsection{Retry data improves SQL generation metrics}
\label{paragraph: res_pretrain_retry}

To test if the considered models can learn the ability to self-correct, we continued the pre-training process with reasoning-enhanced training data, both error-free and with \textit{retry data}. Compared to the zero-shot baselines, models continuously pre-trained with error-free data yielded impressive improvements in generation accuracy metrics (\textbf{Tab.~\ref{tab:results_pretraining_best}}).
These results are expected, as during training we provided the model with previously unseen domain-specific text-to-SQL training samples.

The \textit{retry data} results show us that models continuously pre-trained with such corrupted samples lead to improved accuracy metrics when compared to the error-free continued pre-training (\textbf{Tab.~\ref{tab:results_pretraining_best}}, \textbf{Tab.~S2}). Out of four approaches to \textit{retry data} preparation, the \textit{FS} variant proved to be the best overall. The highest improvements in overall generation accuracy were observed for $p_{retry}$=0.2 and 0.3. For the results pertaining to the other variants, see \textbf{Appendix A.3}. 

The effectiveness of \textit{RetrySQL} transfers across models and benchmark datasets. For OpenCoder~1.5B, the overall Execution Accuracy improved by $\sim$4~p.p. and and 3.1~p.p., measured on BIRD and SPIDER benchmarks, respectively. For Qwen2.5-Coder~1.5B, the overall improvements were 0.4 p.p. and 3.93 p.p., on BIRD and SPIDER, respectively.

While the EX$_{overall}$ metric does show the effectiveness of using \textit{retry data} for improving text-to-SQL generation accuracy in general, it does not show the full picture. \textit{RetrySQL} works particularly well for the hardest evaluation samples (\textbf{Tab.~\ref{tab:results_pretraining_best}}).
Previous research postulated that the advantage of the self-correction ability can be observed especially for complex out-of-distribution evaluation examples, which require the longest solution reasoning sequences~\cite{ye_physics_part2point2_2024}. We found this to be the case for most model and benchmark combinations. It is worth noting that the number of operations necessary to explain SQL queries is not perfectly correlated with the difficulty level (\textbf{Fig.~S1}). As such, many challenging examples can be solved with relatively short queries, which do not require the model to generate long reasoning step sequences. 
This might explain why for OpenCoder~1.5B and BIRD the results are spread more evenly across difficulty levels.

\begin{figure}[t]
    \centering
    \includegraphics[width=0.9\linewidth]{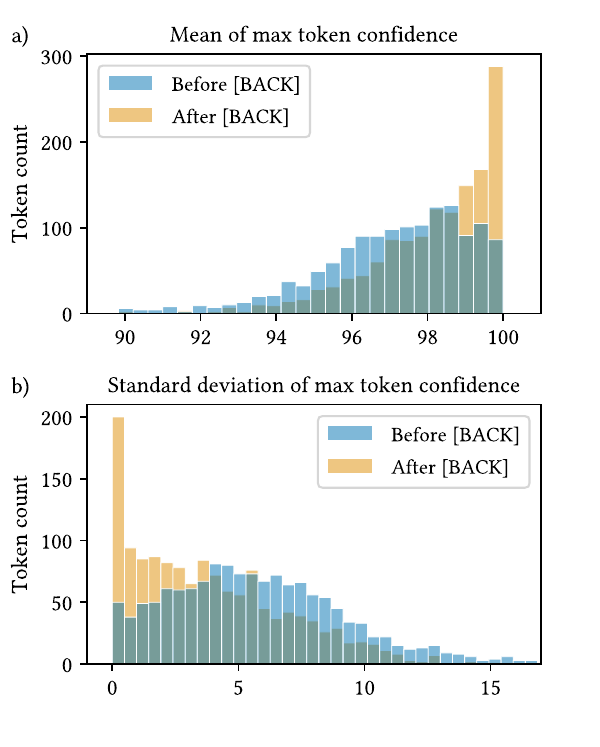}
    \caption{Distribution of token confidence before and after \textbf{[BACK]} tokens. \textbf{(a)} Mean of max token confidence across 10 beam search passes. It can be seen that the confidence score is on average much higher for tokens after the \textbf{[BACK]} token, indicating that the model is uncertain as it makes mistakes, but is confident after self-correction. \textbf{(b)} Standard deviation of max token confidence across 10 beam search passes. The variance of model predictions is much higher as it makes mistakes than after self-correction. 
    }
    \label{fig:token_proba}
\end{figure}

\subsection{Does \textit{RetrySQL} know that it makes mistakes?}
\label{paragraph: res_confidence}

The EX results show us that training with \textit{RetrySQL} increases the number of correct SQL queries compared to models trained without \textit{retry data} (\textbf{Tab.~\ref{tab:results_pretraining_best}}). However, it could be the case that the additional tokens present in reasoning steps in \textit{retry data} simply act as robust augmentation and the model does not learn to self-correct. The underlying model behavior needs to be studied in more detail. To this end, we analyzed the confidence scores returned by the model trained with the Retry \textit{FS} 0.3 dataset 
in proximity (radius of 10 tokens) to \textbf{[BACK]} tokens. We took max softmax scores from these tokens, and then calculated the mean and standard deviation per token across 10 multinomial beam search passes. Finally, we averaged these metrics separately for tokens before and after each \textbf{[BACK]} token.

It is evident that the mean of max confidence scores for predicted tokens differs between tokens preceding the \textbf{[BACK]} token and those after it~(\textbf{Fig.~\ref{fig:token_proba}a}). In other words, as the model is making mistakes, it is less confident in its predictions than after it self-corrects itself and starts to generate correct tokens. This shows that the ability to self-correct is an active part of the generation process.

Similarly, the standard deviation of confidence scores across beam search passes differs between tokens before and those after the \textbf{[BACK]} token~(\textbf{Fig.~\ref{fig:token_proba}b}). For the incorrect tokens, the variance is on average much higher than for the ones after the self-correction. This indicates a reduction in model uncertainty - before the \textbf{[BACK]} token each beam search pass returns significantly different results, the model is not decided what to choose. Conversely, after self-correction, the results become much more consistent and certain - the model catches the error and commits to the correction.

Both of these results show that the self-correcting ability is a learned behavior, resulting from the inclusion of \textit{retry data} in the training process. This new skill is evident when we analyze the generated reasoning steps and SQL queries. The error-free model can hallucinate non-existent tables or JOIN conditions in its reasoning and then include them in the SQL query. The \textit{RetrySQL}-trained model generates correct SQLs in these cases. For examples of such model outputs, see \textbf{Appendix~A.9}. For an additional analysis of generated reasoning steps, see \textbf{Appendix~A.4}.

\subsection{RetrySQL-trained model in an end-to-end text-to-SQL pipeline}
\label{paragraph: res_pipeline}

All experiments presented thus far focused strictly on the generation step, with \textit{perfect} pre-computed schema linking. 
However, in a real text-to-SQL pipeline, perfect schema linking is not available. In order to validate if 
\textit{RetrySQL}-trained models
are competitive with existing proprietary models in the full end-to-end pipeline setting, we performed an additional set of experiments. We observed that the \textit{RetrySQL}-trained models (OpenCoder~1.5B, BIRD training data) compare favorably to much larger models (\textbf{Tab.~S5}): they achieve EX$_{overall}$ of up to 51.36, compared to 32.53 and 54.99 for GPT-4o-mini and GPT-4o, respectively. For the complete description of the pipeline, together with details on our schema linking methodology and full evaluation results, see \textbf{Appendix~A.5}.
These results indicate that using \textit{retry data} in conjunction with our \textit{RetrySQL} method produces 1.5B-parameter models that are competitive with much larger proprietary models ($\sim$8B and $\sim$200B parameters for GPT-4o-mini and GPT-4o~\cite{abacha_medec_2025}, respectively).
This is a promising outcome, showing that incorporating self-correction in the generation stage might be a way forward for future text-to-SQL end-to-end pipelines. 

\section{Limitations}

In our experiments we used the training data from the BIRD and SPIDER benchmark datasets, which contain a limited number of training examples. This is different than what has been done in previous work on self-correction with \textit{retry data}~\cite{ye_physics_part2point2_2024}. There, the training examples were generated on demand as the training went on, to fill a preset number of training steps. We did not have a setup for generating synthetic data in that way, and had to rely on the curated training examples from BIRD. However, a direct comparison to our results is not obvious, as the problem setting of grade school math is very different to our text-to-SQL task. In addition, the metric used in that work was a direct measure of correctness, while in our case we used an indirect Execution Accuracy metric computed in relation to database values, which were not present in the training data. 
We leave synthetic training data generation as a topic for future work.

We did not consider evaluating \textit{RetrySQL} scaling with model size, because we specifically wanted to address the problem of small model performance in the text-to-SQL task. In addition, there are some challenges related to larger models. According to LLM scaling laws~\cite{kaplan_scaling_2020}, in order to avoid overfitting when model size is increased, the data volume has to increase accordingly. As indicated above, we did not have access to large annotated text-to-SQL datasets. Furthermore, full-parameter training of larger models is quite intensive in terms of compute resources.

For the full pipeline experiments, we utilized only a relatively simple LLM-based schema linking stage and did not include a correction stage at the end. This is not an ideal strategy, as there are many optimizations that could be applied to these stages. However, the main part of our research focused on the generation stage and these other elements remained out of scope for us. Moreover, because the \textit{RetrySQL} training paradigm teaches the generation model to \textit{self}-correct, the additional correction step becomes less important. We leave building a fully optimized end-to-end text-to-SQL pipeline, with the most recent approaches to schema linking and query selection, as a topic for future research.

\section{Conclusions}

In this paper, we presented \textit{RetrySQL}, a novel approach to training text-to-SQL generation models. Our solution utilizes reasoning steps with \textit{retry data} in the training examples, which teaches the generation model to self-correct itself as it produces its output. We show that using such data for the continued pre-training of a~coding LLM leads to improved Execution Accuracy metrics when compared to models pre-trained without \textit{retry data}. 
We provide an explainability analysis for our results - we show that as the \textit{RetrySQL} model makes mistakes, it is less confident in its predictions than after it self-corrects itself. Finally, we showcase that incorporating \textit{RetrySQL}-trained 1.5B-parameter models into a relatively simple end-to-end text-to-SQL pipeline produces results that are competitive with much larger closed-source proprietary LLMs such as GPT-4o-mini and GPT-4o, making them much more viable for real-world deployment. We hope that our \textit{RetrySQL} training paradigm will lead to further developments in text-to-SQL models, especially in the context of self-correcting generation.

\bibliography{aaai2026}

\clearpage

\onecolumn
\appendix
\section{Appendix}

\setcounter{table}{0}
\setcounter{figure}{0}
\renewcommand{\thetable}{S\arabic{table}}
\renewcommand{\thefigure}{S\arabic{figure}}

\subsection{Training details}
\label{paragraph: training_details}

All models were trained with NVIDIA A100 80GB, utilizing 2 GPUs. We utilized virtual machines initialized with the following Docker image: \textit{nvidia/cuda:11.8.0-cudnn8-devel-ubuntu20.04}. For dependency versions used in our code, consult the \textit{pyproject.toml} file. During training, the effective batch size equaled 128, due to gradient accumulation being set to 4 and the per-device batch size set to 16. We used dynamic per-batch right padding, with sequence length being padded to a multiple of 8. We utilized the AdamW optimizer with the following hyperparameters (inspired by~\cite{li_codes_2024}): $\beta_1$=0.9, $\beta_2$=0.95, $epsilon$=1e-08, $weight\_decay$=0.1. In all experiments, we trained for 5 epochs, with $learning\_rate$=5.0e-05 and a cosine learning rate schedule. For optimizing the memory usage, we employed the DeepSpeed framework~\cite{rasley_deepspeed_2020} with stage Zero-2. Each training took approximately 4.47 GPU hours.


\subsection{Zero-shot evaluation results}
\label{paragraph: abs_zero_shot}

Below we present results for the initial zero-shot evaluation with the BIRD dataset. For the proprietary models, we used the prompt template and inference configuration that followed the BIRD baselines~\cite{li_can_2023}, including zero-shot CoT and $temperature=0.0$. We set ${max\_output\_tokens=2048}$ to avoid truncated outputs. 

\begin{table*}[h]
\caption{Execution Accuracy for models evaluated in a zero-shot scenario. Models were prompted using the BIRD baseline prompts, as described in \ref{paragraph: baseline_models} and assuming perfect schema linking.}
\label{tab: baselines}
\centering
\begin{tabular}{c|ccc|c}
    \toprule
    Model name & EX$_{simple}$ & EX$_{moderate}$ & EX$_{challenging}$ & EX$_{overall}$  \\
    \midrule
    GPT-4o & 72.37 ± 0.51 & 53.16 ± 1.31 & 43.59 ± 1.79 & 63.73 ± 0.15 \\
    GPT-4o-mini & 47.31 ± 0.4 & 24.31 ± 1.12 & 21.38 ± 1.38 & 37.91 ± 0.25 \\
    Gemini-1.5-pro & 75.59 ± 0.93 & 59.87 ± 0.67 & 59.03 ± 2.32 & 69.27 ± 0.94 \\
    Gemini-1.5-flash & 74.68 ± 1.04 & 59.01 ± 0.18 & 56.14 ± 1.87 & 68.20 ± 0.84 \\
    \midrule
    Llama-3.2-1B & 17.25 ± 0.24 & 3.84 ± 0.42 & 3.45 ± 0.44 & 11.89 ± 0.19 \\
    Qwen2.5-1.5B & 38.7 ± 0.10 & 15.65 ± 0.32 & 8.00 ± 0.55 & 28.83 ± 0.18 \\
    Qwen2.5-Coder-1.5B & 46.77 ± 0.44 & 24.78 ± 0.39 & 11.45 ± 0.94 & 36.78 ± 0.44 \\
    OpenCoder 1.5B & 40.04 ± 0.20 & 16.90 ± 0.48 & 7.45 ± 0.91 & 29.96 ± 0.07 \\
    \bottomrule
\end{tabular}
\end{table*}

\subsection{Full \textit{RetrySQL} results}
\label{paragraph: abs_full_ablation}

In this section, we provide a full evaluation of all \textit{retry data} variants (\textbf{Tab.~\ref{tab:results_pretraining_all}}) for OpenCoder 1.5B and BIRD, as well as several additional analyses for the results.

First, we assessed the impact of adding error-free reasoning steps to BIRD training data. It is evident that they have a significant impact on model performance by themselves - the overall Execution Accuracy increases by over 17~p.p. compared to training without reasoning steps. However, with \textit{RetrySQL} it is possible to increase the model performance even further by enabling the resulting model to self-correct itself during generation.

The full results demonstrate that the \textit{FS} \textit{retry data} variant is the most effective out of the four that were considered in our study. While the EX$_{overall}$ metric for the other variants improves upon the baseline error-free training in most cases, none of the results match the findings for the \textit{FS} variant. This is even more evident for the detailed difficulty metrics, for which the \textit{FM}, \textit{FBS} and \textit{FBM} variants are either worse than the \textit{FS} variant, or worse than the error-free baseline altogether (see especially the EX$_{challenging}$ metric). These results showcase that in the text-to-SQL task, \textit{retry data} in reasoning steps needs to be sampled in the forward direction once per step, since other strategies are to a~large extent not as efficient.

\begin{table*}[h]
\caption{Execution Accuracy for OpenCoder 1.5B continuously pre-trained with \textit{RetrySQL} and evaluated on BIRD, with all \textit{retry data} variants. All results are expressed in percentages. Since we found that the model variance across multinomial beam search passes is relatively low for the \textit{FS} datasets (\textbf{Tab.~\ref{tab:results_pretraining_best}}), we did not calculate standard deviations for the remaining variants. The best results are marked in bold. Results with \textit{retry data} that improve upon the error-free training are indicated with an underline.}
\label{tab:results_pretraining_all}
\setlength\extrarowheight{0.6pt}
\centering
\aboverulesep = 0.7pt
\belowrulesep = 1.2pt
\begin{tabular}{c|ccc|c}
    \toprule
    Dataset variant & EX$_{simple}$ & EX$_{moderate}$ & EX$_{challenging}$ & EX$_{overall}$  \\
    \midrule
    -- (zero-shot) & 40.04 & 16.90 & 7.45 & 29.96 \\
    \midrule
    error-free (no reasoning) & 43.78 & 28.88 & 24.83 & 37.48 \\
    \midrule
    error-free (with reasoning) & 62.70 & 43.53 & 39.45 & 54.71 \\
    \midrule
    Retry \textit{FS} 0.1 & \underline{65.84} & \underline{44.09} & 36.83 & \underline{56.52} \\
    Retry \textit{FS} 0.2 & \underline{\textbf{68.22}} & \underline{\textbf{45.47}} & \underline{40.28} & \underline{\textbf{58.70}} \\
    Retry \textit{FS} 0.3 & \underline{68.00} & \underline{44.91} & \underline{\textbf{43.31}} & \underline{58.68} \\
    Retry \textit{FS} 0.4 & \underline{66.57} & \underline{44.22} & 34.62 & \underline{56.79} \\
    Retry \textit{FS} 0.5 & \underline{66.98} & \underline{44.96} & 37.79 & \underline{57.56} \\
    \midrule
    Retry \textit{FM} 0.1 & \underline{63.68} & \underline{43.97} & 35.86 & \underline{55.08} \\
    Retry \textit{FM} 0.2 & \underline{64.97} & 41.38 & 39.31 & \underline{55.41} \\
    Retry \textit{FM} 0.3 & \underline{66.81} & \underline{44.18} & 37.24 & \underline{57.17} \\
    Retry \textit{FM} 0.4 & \underline{64.76} & 41.81 & 36.55 & \underline{55.15} \\
    Retry \textit{FM} 0.5 & 57.51 & 37.28 & 28.28 & 48.63 \\
    \midrule
    Retry \textit{FBS} 0.1 & \underline{66.70} & 43.53 & 34.48 & \underline{56.65} \\
    Retry \textit{FBS} 0.2 & \underline{66.59} & \underline{44.40} & 34.48 & \underline{56.84} \\
    Retry \textit{FBS} 0.3 & \underline{67.03} & \underline{45.26} & 35.86 & \underline{57.50} \\
    Retry \textit{FBS} 0.4 & \underline{68.32} & 43.10 & 35.86 & \underline{57.63} \\
    Retry \textit{FBS} 0.5 & \underline{66.16} & 42.67 & 35.17 & \underline{56.13} \\
    \midrule
    Retry \textit{FBM} 0.1 & \underline{66.38} & 41.81 & 37.24 & \underline{56.19} \\
    Retry \textit{FBM} 0.2 & \underline{65.95} & \underline{43.97} & 33.79 & \underline{56.26} \\
    Retry \textit{FBM} 0.3 & \underline{67.46} & 43.53 & 37.93 & \underline{57.43} \\
    Retry \textit{FBM} 0.4 & \underline{66.05} & \underline{44.18} & 31.72 & \underline{56.19} \\
    Retry \textit{FBM} 0.5 & \underline{64.65} & 40.09 & 30.34 & 53.98 \\
    \bottomrule
\end{tabular}
\end{table*}

\begin{figure}[h!]
    \centering
    \includegraphics[width=0.5\linewidth]{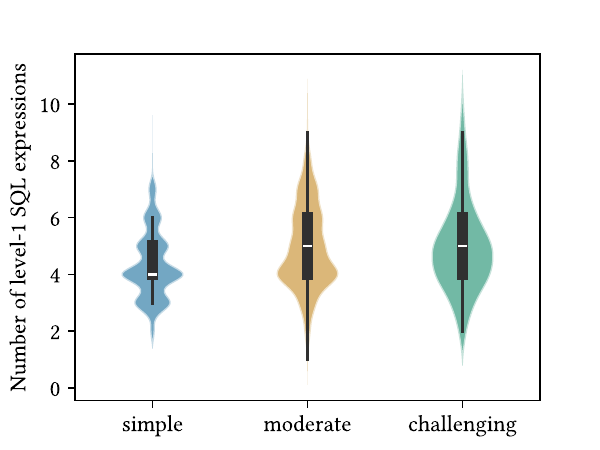}
    \caption{Distribution of SQL query complexity in the BIRD development dataset. There is a significant overlap in the number of level-1 expressions in the SQL syntax tree across difficulty levels defined in the BIRD development dataset. Due to our reasoning generation strategy (see \textbf{Section~\ref{paragraph: reasoning_data}}), the number of these expressions is a proxy for the number of reasoning steps. We parsed the ground truth SQL queries with the SQLGlot Python library~\cite{sqlglot} and extracted level-1 elements from the corresponding syntax trees.}
    \label{fig:difficulty_breakdown}
\end{figure}

\subsection{Additional result analysis}
\label{paragraph: abs_additional_analysis}

We performed several additional analyses to explain the behavior of \textit{RetrySQL}. You can find their outcomes below:

\begin{itemize}
\item \textbf{Query complexity} We plotted the distribution of SQL query complexity across data instances in the BIRD development dataset (\textbf{Fig.~\ref{fig:difficulty_breakdown}}). There is a significant overlap of query complexity for all difficulty levels.
\item \textbf{Number of [BACK] tokens per difficulty} We measured the mean number of generated \textbf{[BACK]} tokens per difficulty level for responses generated by a \textit{RetrySQL}-trained model (OpenCoder 1.5B, Retry \textit{FS} 0.2, BIRD dataset): simple 1.15 ± 0.35, moderate 1.46 ± 1.02, challenging 1.57 ± 0.73. There is no clear correlation in the mean number of \textbf{[BACK]} tokens per difficulty level.
\item \textbf{Reasoning chain length for error-free and \textit{RetrySQL}-trained models} The average number of steps for the error-free model is 12.48 ± 7.73, while for a model trained with Retry \textit{FS} 0.3 it is 9.472 ± 4.18 (\textbf{Tab.~\ref{tab:abs_reasoning_length}}). The reason for the lower number of steps with the \textit{RetrySQL}-trained model comes from the presence of \textbf{[BACK]} tokens, i.e. from the self-correcting ability. The error-free model doesn’t have any means to self-correct, but it knows when a mistake was made (see \textbf{Section~\ref{paragraph: res_linear_probing}}). The only way for it to attempt a correction is to generate more reasoning steps (\textbf{Fig.~\ref{fig:abs_reasoning_length_comparison_example}}). On the other hand, the \textit{RetrySQL}-trained model does have the self-correcting ability, so whenever it detects a mistake, it simply generates a \textbf{[BACK]} token and corrects it, which leads to an overall reduction in the number of required reasoning steps.
\end{itemize}

\begin{table}[h]
    \caption{Mean number of reasoning steps generated by OpenCoder 1.5B. We measured the number of reasoning steps for examples from the BIRD evaluation dataset. Error-free model was trained without \textit{retry data}, while \textit{RetrySQL} was trained with Retry \textit{FS} 0.3.}
    \centering
    \begin{tabular}{c|ccc|c}
        \toprule
         Model variant & Simple & Moderate & Challenging & Overall \\
         \midrule
         \multicolumn{5}{c}{\textbf{All evaluation examples}}\\
         \midrule
         error-free & 11.71 ± 7.67 & 11.82 ± 7.29 & 13.62 ± 8.88 & 12.48 ± 7.73 \\
         \textit{RetrySQL} & 8.638 ± 4.17 & 9.91 ± 4.40 & 9.754 ± 2.97 & 9.472 ± 4.18 \\
         \midrule
         \multicolumn{5}{c}{\textbf{Evaluation examples with generated [BACK] token}}\\
         \midrule
         error-free & 14.516 ± 9.77 & 13.26 ± 8.62 & 16.29 ± 10.52 & 14.34 ± 9.57 \\
         \textit{RetrySQL} & 10.81 ± 5.19 & 12.556 ± 6.03 & 10.57 ± 1.59 & 11.39 ± 5.20 \\
         \midrule
         \multicolumn{5}{c}{\textbf{Evaluation examples without generated [BACK] token}}\\
         \midrule
         error-free & 10.93 ± 6.79 & 11.06 ± 6.38 & 12.23 ± 6.79 & 11.06 ± 6.69 \\
         \textit{RetrySQL} & 8.04 ± 3.62 & 8.55 ± 2.34 & 9.59 ± 3.78 & 8.30 ± 3.37 \\
         \bottomrule
    \end{tabular}
    \label{tab:abs_reasoning_length}
\end{table}



\subsection{Text-to-SQL pipeline}
\label{paragraph: abs_pipeline}

In this section we describe our full text-to-SQL pipeline and provide more insights into all ablations and experiments conducted during the design process for the best schema linking approach.

\subsubsection{Pipeline description}

Our full text-to-SQL pipeline consisted of two modules:

\begin{itemize}
    \item \textbf{schema linking}: an LLM-based schema linker was executed in order to connect the correct tables and columns with the natural language question, thus retrieving the essential context for the generation step;
    \item \textbf{generation}: \textit{RetrySQL-FS-0.2} and \textit{RetrySQL-FS-0.3} models were used to generate a sequence of reasoning steps and the final SQL query.
\end{itemize}



\subsubsection{Schema linking}

Schema linking is one of the most critical steps within the full text-to-SQL pipeline. While recent advancements suggest that it might be omitted in the case of using LLMs for the query generation~\cite{maamari_death_2024}, feeding the model with too much data can provoke hallucinations and increase the inference cost due to a large number of processed input tokens~\cite{chung_longcontext_2025}. Moreover, the full database schema might simply not fit into the context window of the model.

Inspired by leading solutions in the BIRD benchmark~\cite{pourreza_din-sql_2023, pourreza_dts-sql_2024}, we treated schema linking as a separate task and designed a bespoke experimentation framework solely for the purpose of evaluating selected schema linking solutions. We extracted ground truth schema links from the BIRD development set and then investigated several schema linking algorithms. The tested algorithms can be grouped in the following categories:

\begin{itemize}
    \item \textbf{heuristic based methods}: algorithms that do not use any machine learning (ML) techniques, and instead try to link table and column names either by exact matching or by employing edit distance thresholds;
    \item \textbf{embedding based methods}: ML-based algorithms that embed column and table representations and try to match them in the vector space via embedding similarity;
    \item \textbf{LLM-based methods}: LLM-based algorithms that aim to find the correct schema linking by prompting an LLM to solve this specific task.
\end{itemize}

We compared these approaches by measuring the following metrics:
\begin{itemize}
    \item FP - false positive rate, indicates the proportion of irrelevant columns retrieved over the total number of columns;
    \item recall$_{col}$ - fraction of correctly retrieved columns per example, averaged over all test examples;
    \item recall$_{link}$ - proportion of test examples for which all required columns for perfect schema linking were retrieved.
\end{itemize}

Both embedding-based and LLM-based methods were evaluated in zero-shot mode (no training or fine-tuning involved), using the BIRD development set. For embedding-based models, we performed k-NN search with \textit{k}=30. We prepared the queries by joining the natural language question and external knowledge, while documents were represented and stored as the set of all possible combinations of \textit{table name - column name - column description - column data format} in the target database.

Our results show that LLM-based schema linking approaches represent the current best method for solving the schema linking problem~(\textbf{Tab.~\ref{tab: schema_linking}}). Among the tested cloud-based LLMs, Gemini-1.5-pro achieved the best results: 0.37 FP, 0.92 recall$_{col}$, 0.77 recall$_{link}$. Thus, Gemini-1.5-pro was chosen for the schema linking stage of our pipeline. Empirically, we observed that specifying foreign and primary key metadata in the database schema boosted the results for all implemented methods. For completeness, we provide the prompt used for retrieving the schema linking and an example of the model's output (\textbf{Fig.~\ref{fig:schema_linking_prompt}}).

\begin{table*}[t]
\caption{Results for the schema linking experiments. The best performing approach, as measured by all metrics, was the LLM-based schema linking with Gemini-1.5-pro. For FP, lower values are better. For recall$_{col}$ and recall$_{link}$ higher values are better. The best results are marked in bold.}
\label{tab: schema_linking}
\centering
\begin{tabular}{c|ccc}
    \toprule
    Method & FP & recall$_{col}$ & recall$_{link}$  \\
    \midrule
        \multicolumn{4}{c}{\textbf{Heuristic-based methods}} \\
    \midrule
    exact-matching & 0.75 & 0.54 & 0.16\\
    edit-distance=1 & 0.79 & 0.62 & 0.25\\
    edit-distance=2 & 0.84 & 0.66 & 0.27\\
    edit-distance=3 & 0.9 & 0.72 & 0.33\\
    \midrule
        \multicolumn{4}{c}{\textbf{Embedding-based methods}} \\
    \midrule
    OpenAI - text-embeddings-ada-002 & 0.73 & 0.64 & 0.28\\
    OpenAI - text-embeddings-3-large & 0.88 & 0.81 & 0.54\\
    Snowflake - arctic-embed-m \cite{arctic_embed} & 0.75 & 0.6 & 0.23\\
    Snowflake - arctic-embed-l \cite{arctic_embed} & 0.89 & 0.78 & 0.48\\
    \midrule
        \multicolumn{4}{c}{\textbf{LLM-based methods}} \\
    \midrule
    GPT-4o & 0.38 & 0.92 & 0.76\\
    GPT-4o-mini & 0.43 & 0.85 & 0.56\\
    Gemini-1.5-pro & \textbf{0.37} & \textbf{0.92} & \textbf{0.77}\\
    Gemini-1.5-flash & 0.39 & 0.91 & 0.72\\
    \bottomrule
\end{tabular}
\end{table*}

\subsubsection{Pipeline results}
Here we provide Execution Accuracy results for the full pipeline setting, evaluated with the BIRD benchmark dataset. \textit{RetrySQL}-trained OpenCoder~1.5 models compare favorably to much larger, proprietary models: they achieve EX$_{overall}$ of up to 51.36, compared to 32.53 and 54.99 for GPT-4o-mini and GPT-4o. Crucially, they still improve over error-free training in this case as well, which shows that \textit{RetrySQL} remains effective even with non-perfect schema linking.

\begin{table*}[h]
\caption{Execution Accuracy for the full end-to-end text-to-SQL pipeline. All results are expressed in percentages, with mean and standard deviation over 5 multinomial beam search generations.}
\label{tab:results_pipeline}
\centering
\begin{tabular}{c|ccc|c}
    \toprule
    Model name & EX$_{simple}$ & EX$_{moderate}$ & EX$_{challenging}$ & EX$_{overall}$  \\
    \midrule
    GPT-4o & 61.62 ± 3.83 & 42.74 ± 2.34 & 40.48 ± 1.78 & 54.99 ± 0.32 \\
    GPT-4o-mini & 42.12 ± 1.06 & 18.06 ± 0.47 & 17.65 ± 0.38 & 32.53 ± 0.72 \\
    Gemini-1.5-pro & 66.88 ± 0.23 & 48.15 ± 0.62 & 51.86 ± 0.58 & 59.79 ± 0.31 \\
    Gemini-1.5-flash & 64.69 ± 0.12 & 45.86 ± 0.24 & 49.52 ± 0.57 & 57.56 ± 0.08 \\
    \midrule
    error-free & 55.95 ± 0.09 & 38.94 ± 0.09 & 31.03 ± 0.0 & 48.47 ± 0.07 \\
    \midrule 
    \textit{RetrySQL-FS-0.2} & 59.81 ± 0.04 & 37.46 ± 0.09 & 35.17 ± 0.00 & 50.72 ± 0.00 \\
    \textit{RetrySQL-FS-0.3} & 60.28 ± 0.05 & 38.36 ± 0.14 & 36.00 ± 0.28 & 51.36 ± 0.05 \\
    \bottomrule
\end{tabular}
\end{table*}

\subsection{Linear probing experiments}
\label{paragraph: abs_linear_probing_exps}
In this section we provide more details regarding our linear probing experiments, which were inspired by a probing task originally introduced in previous work~\cite{ye_physics_part2point1_2024} (specifically, the $can\_next(A)$ task).




\subsubsection{Experiment setup}
We prepared the linear probing model by taking the OpenCoder-1.5B weights continuously pre-trained with error-free reasoning steps and replacing the existing head with a linear one. The new head mapped the $2044$-dimensional vector of the last token in the input sequence to a single sigmoid-activated neuron for the binary classification task. Unlike~\cite{ye_physics_part2point1_2024}, we did not introduce any small rank-r update on the input (embedding) layer. We trained this classification model using a machine equipped with 2 x NVIDIA A100 80GB, with effective $batch\_size$=128 and $learning\_rate$=1e-4. We used the AdamW optimizer with $\beta_1$=0.9, $\beta_2$=0.95 and utilized \textit{early stopping}. 

We trained the model in a 5-fold cross-validation setup, by retaining $80\%$ of the dataset for training and using the rest for validation. We used the best checkpoints to compute $balanced\_accuracy$ and $f1\_score$ (reported for the \textit{incorrect} class).

\subsubsection{Discussion}
By performing the linear probing experiment we demonstrated that a coding-oriented LLM, previously pre-trained with only error-free reasoning steps, exhibits \textit{regretful} patterns during inference for input examples that contain an \textit{incorrect} sequence of reasoning steps. 
It is possible to leverage the probing results by applying the classifier described above to guide the SQL generation process. After generating each solution sentence, the next probing could be used to determine whether the model knows that it has made a mistake and, if so, it could be reverted to the end of the previous sentence and regenerate from that point. Previous work showed that this method can increase generation accuracy in mathematical reasoning tasks, at the cost of increased inference complexity~\cite{ye_physics_part2point2_2024}. We did not perform such an analysis, since we treated the probing experiment as a preliminary sanity check for using \textit{retry data} in the text-to-SQL task.

\clearpage

\subsection{Training data example}

\begin{figure}[h]
\centering
\begin{tikzpicture}
    \node[draw, fill=gray!20, 
          rectangle, 
          rounded corners=10pt, 
          minimum width=13.5cm, 
          minimum height=8.1cm, 
          text width=11cm,
          align=left, 
          drop shadow] 
    {   
        \textbf{[CONTEXT]} \\\texttt{CREATE TABLE 'games' \\
        (\\
        \quad 'games\_year' INTEGER DEFAULT null \\
        );} \\

        -{}- External Knowledge: games refer to id; during the 90s refers to \par
        \texttt{games\_year BETWEEN 1990 AND 1999;} \par
        -{}- Using valid SQLite and understanding External Knowledge, answer the following questions for the tables provided above. \\[1em]

        \textbf{[QUESTION]} How many Olympics games were held during the 90s? \\[1em]
        \textbf{[REASONING]} Define the main table in the \texttt{FROM} clause: \texttt{FROM games}. \par
        Define the filtering condition in the \texttt{WHERE} clause: \texttt{WHERE games\_year BETWEEN 1990 AND 1999}. \par
        Select the column to be included in the final result: \texttt{SELECT COUNT(games\_year)}.\\[1em]

        \textbf{[SQL]} \texttt{SELECT COUNT(games\_year) FROM games WHERE games\_year BETWEEN '1990' AND '1999';}
    };
\end{tikzpicture}
\caption{Example of a training data sample used in our experiments. It consists of the following elements: DDL statements for schema representation, external knowledge and question extracted from the BIRD metadata, reasoning steps generated as described in Section \ref{paragraph: reasoning_data} and the ground truth SQL query. These components are separated by special tokens to guide the model in the learning process.}
\label{fig:training_sample}
\end{figure}

\newpage

\subsection{Prompts}
\label{paragraph: prompts}

\begin{figure*}[h]
\centering
\scalebox{.85}{
\begin{tikzpicture}
    \node[draw, fill=yellow!30, 
          rectangle, 
          rounded corners=10pt, 
          minimum width=17cm, 
          minimum height=20cm, 
          text width=15.5cm, 
          align=left, 
          drop shadow] 
    {   \small
        You are an SQL expert. When I provide you with a SQL query, your task is to describe step-by-step how a person would create such a query. Follow the standard SQL execution order. Write the steps from the perspective of someone constructing the query. If the query includes subqueries, describe them step-by-step in the same detailed manner as the main query before referencing them. Each step should represent a distinct operation. Make the operations as granular as possible. \\[1em]
        
        Here is the standard SQL execution order you should follow for your explanation. If a given clause is not present in the query, skip it without mentioning its absence: \\
        1. FROM clause (including JOINs). \\
        2. WHERE clause. \\
        3. GROUP BY clause. \\
        4. HAVING clause. \\
        5. SELECT clause. \\
        6. ORDER BY clause. \\
        7. LIMIT clause. \\[1em]
        
        For each query I provide: \\
        1. Explain the query step by step in plain language. \\
        2. Ensure that each step corresponds to one small, logical operation. \\
        3. Use clear and concise language for each operation. \\
        4. Each step should be provided in a single line (use single newline character between steps). \\[1em]

        Here is an example query for your reference: \\
        \texttt{SELECT T1.name, T1.email, SUM(T3.amount) AS total\_sales FROM Customers AS T1 INNER JOIN Orders AS T2 ON T1.customer\_id = T2.customer\_id LEFT JOIN OrderDetails AS T3 ON T2.order\_id = T3.order\_id WHERE T2.order\_date >= '2023-01-01' AND T2.order\_date <= (SELECT order\_date FROM Orders WHERE order\_id = '1' ORDER BY order\_date DESC LIMIT 1) GROUP BY T1.name, T1.email ORDER BY total\_sales DESC} \\[1em]
        
        The expected step-by-step breakdown for the above query: \\
        Define the main table in the FROM clause: \texttt{FROM Customers AS T1}. \\
        Define the first JOIN operation: \texttt{INNER JOIN}. \\
        Define the table to join: \texttt{Orders AS T2}. \\
        Define the join condition: \texttt{ON T1.customer\_id = T2.customer\_id}. \\
        Define the second JOIN operation: \texttt{LEFT JOIN}. \\
        Define the table to join: \texttt{OrderDetails AS T3}. \\
        Define the join condition: \texttt{ON T2.order\_id = T3.order\_id}. \\
        Define the main filtering condition in the WHERE clause: \texttt{WHERE T2.order\_date >= '2023-01-01'}. \\
        Add the additional filtering condition in the WHERE clause : \texttt{AND T2.order\_date <= (subquery)}. \\
        Define the main table in the subquery's FROM clause: \texttt{FROM Orders}. \\
        Define the main filtering condition in the subquery's WHERE clause: \texttt{WHERE order\_id = '1'}. \\
        Select the column to be included in the subquery result: \texttt{SELECT order\_date}. \\
        Order the subquery results by the specified column: \texttt{ORDER BY order\_date DESC}. \\
        Limit the subquery results: \texttt{LIMIT 1}. \\
        Complete the filtering condition in the WHERE clause: \texttt{AND T2.order\_date <= (SELECT order\_date FROM Orders WHERE order\_id = '1' ORDER BY order\_date DESC LIMIT 1)}. \\
        Group the results by the specified columns: \texttt{GROUP BY T1.name, T1.email}. \\
        Select the columns to be included in the final result: \texttt{SELECT T1.name, T1.email, SUM(T3.amount) AS total\_sales}. \\
        Order the results by the specified column: \texttt{ORDER BY total\_sales DESC}. \\[1em]

        Now, I will provide you with a query, and I expect you to respond in this format: \\[1em]

        ${\{sql\_query\}}$ \\
    };
\end{tikzpicture}
}
    \caption{Prompt used for generating reasoning steps.}
    \label{fig:reasoning_prompt}
\end{figure*}

\begin{figure*}[t]
\centering
\begin{tikzpicture}
\node[draw, fill=gray!10,
          rectangle,
          rounded corners=10pt,
          minimum width=17cm,
          minimum height=17.7cm,
          text width=15cm,
          align=left,
          drop shadow]
{
};
\node at (0, 3.15) [draw, fill=yellow!30, 
      rectangle, 
      rounded corners=10pt, 
      minimum width=16.3cm, 
      minimum height=10.5cm, 
      text width=15cm,
      align=left] 
{
    -{}- Database Schema: \\[1em]
    \texttt{CREATE TABLE `yearmonth`\\
(\\
\quad `customerid` INTEGER  not null,\\
\quad `date` TEXT  not null,\\
\quad `consumption` REAL  null,\\
\quad FOREIGN KEY (`customerid`) REFERENCES `customers` (`customerid`),\\
\quad FOREIGN KEY (`customerid`) REFERENCES `customers` (`customerid`),\\
\quad PRIMARY KEY (`date`, `customerid`)\\
);\\
CREATE TABLE `customers`\\
(\\
\quad `customerid` INTEGER  not null  PRIMARY KEY,\\
\quad `currency` TEXT  null\\
);}\\[1em]
-{}- External Knowledge: Pays in euro = \texttt{Currency = 'EUR'} \\[1em]

-{}- Based on Database Schema provided above and understanding External Knowledge, your task is to select table-column pairs (called \textit{schema links}) most relevant to the given Question.\\
-{}- Question: Among the customers who paid in euro, how many of them have a monthly consumption of over 1000? \\[1em]
    Choose the relevant table-column pairs after thinking step by step:
};
\node at (0, -5.50) [draw, fill=cyan!20,
      rectangle,
      rounded corners=10pt,
      minimum width=16.3cm,
      minimum height=6cm,
      text width=15cm,
      align=left]
{
\texttt{
\{ \\
\quad "schema\_links": [ \\
\quad\quad \{ \\
\quad\quad\quad "table\_name": "customers", \\
\quad\quad\quad "columns": [ "customerid", "currency" ] \\
\quad\quad \}, \\
\quad\quad \{ \\
\quad\quad\quad "table\_name": "yearmonth", \\
\quad\quad\quad "columns": [ "consumption", "customerid", "date" ] \\
\quad\quad \} \\
\quad ] \\
\}
}
};
\end{tikzpicture}
    \caption{Example of a prompt used for the schema linking step in our proposed text-to-SQL full pipeline. The prompt is marked with yellow background, while the response is denoted with blue background. Database schemas are represented in form of DDL statements as described in Section \ref{paragraph: training_data}. We used the LangChain library to query LLMs, leveraging its Python dataclass serialization support. Note: the injected instructions for the output JSON format are not included in this example.
    }
    \label{fig:schema_linking_prompt}
\end{figure*}

\clearpage
\subsection{RetrySQL generation examples}
\label{paragraph: abs_generation_examples}

\begin{figure*}[h]
\centering
\begin{tikzpicture}
\node[draw, fill=gray!10, 
          rectangle, 
          rounded corners=10pt, 
          minimum width=17cm, 
          minimum height=12.75cm, 
          text width=15cm,
          align=left, 
          drop shadow] 
{
};
\node at (0, 2.37) [draw, fill=yellow!30, 
      rectangle, 
      rounded corners=10pt, 
      minimum width=16.5cm, 
      minimum height=7.5cm, 
      text width=15cm,
      align=left] 
{
\textbf{[CONTEXT]}\\
\texttt{CREATE TABLE `cards`\\
(\\
\quad `id` INTEGER  not null  PRIMARY KEY,\\
\quad `artist` TEXT  null,\\
\quad `cardkingdomfoilid` TEXT  null,\\
\quad `cardkingdomid` TEXT  null\\
);}\\
-{}- External Knowledge: designed by Aaron Miller refers to \texttt{artist = 'Aaron Miller'}; are icredibily powerful refers to \texttt{cardKingdomFoilId is not null AND cardKingdomId is not null}\\
-{}- Using valid SQLite and understanding External Knowledge, answer the following questions for the tables provided above.\\[\baselineskip]
\textbf{[QUESTION]}\\
Of all the cards that are designed by Aaron Miller, how many of them are incredibly powerful?\\[\baselineskip]
\textbf{[REASONING]}
};
\node at (0, -3.87) [draw, fill=cyan!20, 
      rectangle, 
      rounded corners=10pt, 
      minimum width=16.5cm, 
      minimum height=4.5cm, 
      text width=15cm,
      align=left] 
{
Define the main table in the \texttt{FROM} clause: \texttt{`FROM cards`}.\\
\st{Add the additional filtering condition in the \texttt{WHERE} clause: \texttt{`AND cardKingdomFoilId IS NOT NULL`}.} \textbf{[BACK]}\\
Define the main filtering condition in the \texttt{WHERE} clause: \texttt{`WHERE artist = 'Aaron Miller'`}. \\
Add the additional filtering condition in the \texttt{WHERE} clause: \texttt{`AND cardKingdomFoilId IS NOT NULL`}. \\
Select the column to be included in the final result: \texttt{`SELECT COUNT(id)`}. \\[\baselineskip]

\textbf{[SQL]}\\
\texttt{SELECT COUNT(id) FROM cards WHERE artist = 'Aaron Miller' AND cardKingdomFoilId IS NOT NULL AND cardKingdomId IS NOT NULL"}
};
\end{tikzpicture}
\caption{\textit{RetrySQL} generation example with prompt and completion. The prompt is marked with yellow background, while the response is denoted with blue background. The model corrects itself after erroneously generating a reasoning step in its CoT process. Specifically, the model generates an \textit{additional} filtering condition when the \textit{main} one has not been declared yet. After generating the \textbf{[BACK]} token, it \textit{retries} and correctly generates the main filtering condition, and then proceeds until the end of the reasoning chain, finally producing a correct SQL query.}
\label{figure:generation_example_1}
\end{figure*}

\begin{figure*}[h]
\centering
\begin{tikzpicture}
\node[draw, fill=gray!10, 
          rectangle, 
          rounded corners=10pt, 
          minimum width=17cm, 
          minimum height=15.5cm, 
          text width=15cm,
          align=left, 
          drop shadow] 
{
};
\node at (0, 3.75) [draw, fill=yellow!30, 
      rectangle, 
      rounded corners=10pt, 
      minimum width=16.5cm, 
      minimum height=7.5cm, 
      text width=15cm,
      align=left] 
{
\textbf{[CONTEXT]}\\
\texttt{CREATE TABLE `cards`\\
(\\
\quad `id` INTEGER  not null  PRIMARY KEY,\\
\quad `artist` TEXT  null,\\
\quad `convertedmanacost` REAL  null,\\
\quad `side` TEXT  null\\
);}\\
-{}- External Knowledge: doesn't have multiple faces refers to \texttt{side IS NULL}; illustrator refers to \texttt{artist}\\
-{}- Using valid SQLite and understanding External Knowledge, answer the following questions for the tables provided above.\\[\baselineskip]
\textbf{[QUESTION]}\\
Among the cards that doesn't have multiple faces on the same card, who is the illustrator of the card art that has the highest cost of converted mana?\\[\baselineskip]
\textbf{[REASONING]}
};
\node at (0, -3.87) [draw, fill=cyan!20, 
      rectangle, 
      rounded corners=10pt, 
      minimum width=16.5cm, 
      minimum height=7.25cm, 
      text width=15cm,
      align=left] 
{
Define the main table in the \texttt{FROM} clause: \texttt{`FROM cards`}. \\
Define the main filtering condition in the \texttt{WHERE} clause: \texttt{`WHERE side IS NULL`}. \\
Select the column to be included in the final result: \texttt{`SELECT artist`}. \\
\st{Limit the results: \texttt{`LIMIT 1`}.} \textbf{[BACK]} \\
Add the additional filtering condition in the \texttt{WHERE} clause: \texttt{`AND convertedManaCost = (subquery)`}. \\
Define the main table in the subquery's \texttt{FROM} clause: \texttt{`FROM cards`}. \\
Complete the filtering condition in the \texttt{WHERE} clause: \texttt{`AND convertedManaCost = (SELECT MAX(convertedManaCost) FROM cards)`}. \\
Select the column to be included in the final result: \texttt{`SELECT artist`}. \\
\st{Limit the results: \texttt{`LIMIT 1`}.} \textbf{[BACK]} \\
Order the results by the specified column: \texttt{`ORDER BY artist DESC`}. \\
Limit the results: \texttt{`LIMIT 1`}. \\[\baselineskip]

\textbf{[SQL]}\\
\texttt{SELECT artist FROM cards WHERE side IS NULL AND convertedManaCost = ( SELECT MAX(convertedManaCost) FROM cards ) ORDER BY artist DESC LIMIT 1}
};
\end{tikzpicture}
\caption{\textit{RetrySQL} generation example with prompt and completion. The prompt is marked with yellow background, while the response is denoted with blue background. The model corrects itself after erroneously trying to halt the reasoning chain too early. This happens two times along the process. The \textit{retry} mechanism allows the model to understand the mistakes and correct itself. Specifically, the model corrects itself by adding additional \texttt{WHERE} and \texttt{ORDER} operations. After each correction, the model proceeds to generate the rest of the reasoning chain, finally producing a correct SQL query.}
\label{figure:generation_example_2}
\end{figure*}

\begin{figure*}[h]
\centering
\begin{tikzpicture}
\node[draw, fill=gray!10, 
          rectangle, 
          rounded corners=10pt, 
          minimum width=17cm, 
          minimum height=17cm, 
          text width=15cm,
          align=left, 
          drop shadow] 
{
};
\node at (0, 3.37) [draw, fill=yellow!30, 
      rectangle, 
      rounded corners=10pt, 
      minimum width=16.5cm, 
      minimum height=9.75cm, 
      text width=15cm,
      align=left] 
{
\textbf{[CONTEXT]}\\
\texttt{CREATE TABLE `yearmonth`\\
(\\
\quad `customerid` INTEGER  not null,\\
\quad `date` TEXT  not null,\\
\quad `consumption` REAL  null,\\
\quad FOREIGN KEY (`customerid`) REFERENCES `customers` (`customerid`),\\
\quad FOREIGN KEY (`customerid`) REFERENCES `customers` (`customerid`),\\
\quad PRIMARY KEY (`date`, `customerid`)\\
);\\
CREATE TABLE `customers`\\
(\\
\quad `customerid` INTEGER  not null  PRIMARY KEY,\\
\quad `currency` TEXT  null\\
);}\\
-{}- External Knowledge: Pays in euro = \texttt{Currency = 'EUR'}.\\
-{}- Using valid SQLite and understanding External Knowledge, answer the following questions for the tables provided above.\\[\baselineskip]
\textbf{[QUESTION]}\\
Among the customers who paid in euro, how many of them have a monthly consumption of over 1000?\\[\baselineskip]
\textbf{[REASONING]}
};
\node at (0, -5) [draw, fill=cyan!20, 
      rectangle, 
      rounded corners=10pt, 
      minimum width=16.5cm, 
      minimum height=6.5cm, 
      text width=15cm,
      align=left] 
{
Define the main table in the \texttt{FROM} clause: \texttt{`FROM customers AS T1`}.\\
Define the \texttt{JOIN} operation: \texttt{`INNER JOIN`}.\\
Define the table to join: \texttt{`yearmonth AS T2`}.\\
\st{Select the column to be included in the final result: \texttt{`SELECT COUNT(T1.CustomerID)`}.} \textbf{[BACK]}\\
Define the join condition: \texttt{`ON T1.CustomerID = T2.CustomerID`}.\\
\st{Select the column to be included in the final result: \texttt{`SELECT COUNT(T1.CustomerID)`}.} \textbf{[BACK]}\\
Define the main filtering condition in the \texttt{WHERE} clause: \texttt{`WHERE T2.Consumption > 1000`}.\\
\st{Select the column to be included in the final result: \texttt{`SELECT COUNT(T1.CustomerID)`}.} \textbf{[BACK]}\\
Add the additional filtering condition in the \texttt{WHERE} clause: \texttt{`AND T1.Currency = 'EUR'`}.\\
Select the column to be included in the final result: \texttt{`SELECT COUNT(T1.CustomerID)`}.\\[\baselineskip]

\textbf{[SQL]}\\
\texttt{SELECT COUNT(T1.CustomerID) FROM customers AS T1 INNER JOIN yearmonth AS T2 ON T1.CustomerID = T2.CustomerID WHERE T2.Consumption > 1000 AND T1.Currency = 'EUR'}
};
\end{tikzpicture}
\caption{\textit{RetrySQL} generation example with prompt and completion. The prompt is marked with yellow background, while the response is denoted with blue background. The model corrects itself after erroneously trying to halt the reasoning chain too early. This happens three times along the process. The \textit{retry} mechanism allows the model to understand the mistakes and correct itself. In particular, the model corrects itself by adding additional \texttt{JOIN} and \texttt{WHERE} operations. After each correction, the model proceeds to generate the rest of the reasoning chain, finally producing a correct SQL query.}
\label{figure:generation_example_3}
\end{figure*}

\begin{figure*}[h]
\centering
\begin{tikzpicture}
\node[draw, fill=gray!10, 
          rectangle, 
          rounded corners=10pt, 
          minimum width=17cm, 
          minimum height=15.5cm, 
          text width=15cm,
          align=left, 
          drop shadow] 
{
};
\node at (0, 6.5) [draw, fill=yellow!30, 
      rectangle, 
      rounded corners=10pt, 
      minimum width=16.5cm, 
      minimum height=2cm, 
      text width=15cm,
      align=left] 
{
-{}- External Knowledge: \textbf{unknown power refers to \texttt{power = '*'} or \texttt{POWER IS NULL}}; commander play format refers to \texttt{format = 'commander'}; legal for commander play format refers to \texttt{format = 'commander'} where \texttt{status = 'Legal'}
};
\node at (0, 2.25) [draw, fill=pink!20, 
      rectangle, 
      rounded corners=10pt, 
      minimum width=16.5cm, 
      minimum height=6cm, 
      text width=15cm,
      align=left] 
{
Define the main table in the \texttt{FROM} clause: \texttt{`FROM cards AS T1`}.\\
Define the \texttt{JOIN} operation: \texttt{`INNER JOIN`}.\\
Define the table to join: \texttt{`legalities AS T2`}.\\
Define the join condition: \texttt{`ON T1.uuid = T2.uuid`}.\\
Define the main filtering condition in the \texttt{WHERE} clause: \texttt{`WHERE T1.power = '*'`}.\\
Add the additional filtering condition in the \texttt{WHERE} clause: \texttt{`AND T2.format = 'commander'`}.\\
Add the additional filtering condition in the \texttt{WHERE} clause: \texttt{`AND T2.status = 'Legal'`}.\\
Select the columns to be included in the final result: \texttt{`SELECT T1.id, T1.artist`}.\\[\baselineskip]

\textbf{[SQL]}\\
\texttt{SELECT T1.id, T1.artist FROM cards AS T1 INNER JOIN legalities AS T2 ON T1.uuid = T2.uuid WHERE \textbf{\textcolor{orange}{T1.power = '*'}} AND T2.format = 'commander' AND T2.status = 'Legal'}
};
\node at (0, -4.25) [draw, fill=cyan!20, 
      rectangle, 
      rounded corners=10pt, 
      minimum width=16.5cm, 
      minimum height=6.5cm, 
      text width=15cm,
      align=left] 
{
Define the main table in the \texttt{FROM} clause: \texttt{`FROM cards AS T1`}.\\
Define the \texttt{JOIN} operation: \texttt{`INNER JOIN`}.\\
Define the table to join: \texttt{`legalities AS T2`}.\\
Define the join condition: \texttt{`ON T1.uuid = T2.uuid`}.\\
Define the main filtering condition in the \texttt{WHERE} clause: \texttt{`WHERE T1.power = '*'`}.\\
\st{Select the columns to be included in the final result: \texttt{`SELECT T1.id, T1.artist`}.} \textbf{[BACK]}\\
\textbf{\textcolor{green!65!black}{Add the additional filtering condition in the \texttt{WHERE} clause: \texttt{`OR T1.power IS NULL`}.}}\\
\st{Select the columns to be included in the final result: \texttt{`SELECT T1.id, T1.artist`}.} \textbf{[BACK]}\\
Add the additional filtering condition in the \texttt{WHERE} clause: \texttt{`AND T2.status = 'Legal'`}.\\
Select the columns to be included in the final result: \texttt{`SELECT T1.id, T1.artist`}.\\[\baselineskip]

\textbf{[SQL]}\\
\texttt{SELECT T1.id, T1.artist FROM cards AS T1 INNER JOIN legalities AS T2 ON T1.uuid = T2.uuid WHERE \textbf{\textcolor{green!65!black}{T1.power = '*' OR T1.power IS NULL}} AND T2.format = 'commander' AND T2.status = 'Legal'}
};
\end{tikzpicture}
\caption{Comparison of error-free and \textit{RetrySQL} model generation. The external evidence is marked with yellow background, error-free response has a pink background, while the \textit{RetrySQL} response is denoted with blue background. The error-free model omits one condition in the external knowledge (\texttt{POWER IS NULL}). The \textit{RetrySQL}-trained model utilizes the \textit{retry} mechanism to go back from an incorrect step and generate the full condition.}
\label{figure:generation_example_4}
\end{figure*}

\begin{figure*}[h]
\centering
\begin{tikzpicture}
\node[draw, fill=gray!10, 
          rectangle, 
          rounded corners=10pt, 
          minimum width=17cm, 
          minimum height=16.75cm, 
          text width=15cm,
          align=left, 
          drop shadow] 
{
};
\node at (0, 7.37) [draw, fill=yellow!30, 
      rectangle, 
      rounded corners=10pt, 
      minimum width=16.5cm, 
      minimum height=1.5cm, 
      text width=15cm,
      align=left] 
{
-{}- External Knowledge: \textbf{triple bond refers to \texttt{bond\_type = '\#'}}; phosphorus refers to \texttt{element = 'p'}; bromine refers to \texttt{element = 'br'}
};
\node at (0, 2.75) [draw, fill=pink!20, 
      rectangle, 
      rounded corners=10pt, 
      minimum width=16.5cm, 
      minimum height=7.25cm, 
      text width=15cm,
      align=left] 
{
Define the main table in the \texttt{FROM} clause: \texttt{`FROM atom AS T1`}.\\
Define the first \texttt{JOIN} operation: \texttt{`INNER JOIN`}.\\
Define the table to join: \texttt{`bond AS T2`}.\\
Define the join condition: \texttt{`ON T1.molecule\_id = T2.molecule\_id`}.\\
Define the second \texttt{JOIN} operation: \texttt{`INNER JOIN`}.\\
Define the table to join: \texttt{`molecule AS T3`.}\\
Define the join condition: \texttt{`ON T2.molecule\_id = T3.molecule\_id`}.\\
\textbf{\textcolor{red}{Define the main filtering condition in the \texttt{WHERE} clause: \texttt{`WHERE T2.bond\_type = 'triple'`}.}}\\
Add the additional filtering condition in the \texttt{WHERE} clause: \texttt{`AND (T1.element = 'p' OR T1.element = 'br')`}.\\
Select the column to be included in the final result: \texttt{`SELECT COUNT(T1.atom\_id)`}.\\[\baselineskip]

\textbf{[SQL]}\\
\texttt{SELECT COUNT(T1.atom\_id) FROM atom AS T1 INNER JOIN bond AS T2 ON T1.molecule\_id = T2.molecule\_id INNER JOIN molecule AS T3 ON T2.molecule\_id = T3.molecule\_id WHERE \textbf{\textcolor{red}{T2.bond\_type = 'triple'}} AND (T1.element = 'p' OR T1.element = 'br')}
};
\node at (0, -4.63) [draw, fill=cyan!20, 
      rectangle, 
      rounded corners=10pt, 
      minimum width=16.5cm, 
      minimum height=7cm, 
      text width=15cm,
      align=left] 
{
Define the main table in the \texttt{FROM} clause: \texttt{`FROM Molecule AS T1`}.\\
Define the first \texttt{JOIN} operation: \texttt{`INNER JOIN`}.\\
Define the table to join: \texttt{`Atom AS T2`}.\\
Define the join condition: \texttt{`ON T1.molecule\_id = T2.molecule\_id`}.\\
Define the second \texttt{JOIN} operation: \texttt{`INNER JOIN`}.\\
Define the table to join: \texttt{`Bond AS T3`}.\\
Define the join condition: \texttt{`ON T1.molecule\_id = T3.molecule\_id`}.\\
Define the main filtering condition in the \texttt{WHERE} clause: \texttt{`WHERE T2.element IN ('p', 'br')`}.\\
\st{Select the column to be included in the final result: \texttt{`SELECT COUNT(T2.atom\_id)`}.} \textbf{[BACK]}\\
\textbf{\textcolor{green!65!black}{Add the additional filtering condition in the \texttt{WHERE} clause: \texttt{`AND T3.bond\_type = '\#'`}.}}\\
Select the column to be included in the final result: \texttt{`SELECT COUNT(T2.atom\_id)`}.\\[\baselineskip]

\textbf{[SQL]}\\
\texttt{SELECT COUNT(T2.atom\_id) FROM Molecule AS T1 INNER JOIN Atom AS T2 ON T1.molecule\_id = T2.molecule\_id INNER JOIN Bond AS T3 ON T1.molecule\_id = T3.molecule\_id WHERE T2.element IN ('p', 'br') AND \textbf{\textcolor{green!65!black}{T3.bond\_type = '\#'}}}
};
\end{tikzpicture}
\caption{Comparison of error-free and \textit{RetrySQL} model generation. The external evidence is marked with yellow background, error-free response has a pink background, while the \textit{RetrySQL} response is denoted with blue background. The error-free model hallucinates \texttt{bond\_type = `triple`}. The \textit{RetrySQL}-trained model utilizes the \textit{retry} mechanism to go back from an incorrect step and generate the correct condition: \texttt{bond\_type = `\#`}.}
\label{figure:generation_example_5}
\end{figure*}

\begin{figure*}[h]
\centering
\begin{tikzpicture}
\node[draw, fill=gray!10, 
          rectangle, 
          rounded corners=10pt, 
          minimum width=17cm, 
          minimum height=13cm, 
          text width=15cm,
          align=left, 
          drop shadow] 
{
};
\node at (0, 5.5) [draw, fill=yellow!30, 
      rectangle, 
      rounded corners=10pt, 
      minimum width=16.5cm, 
      minimum height=1.5cm, 
      text width=15cm,
      align=left] 
{
-{}- External Knowledge: known as Archenemy refers to \texttt{mcmName = 'Archenemy'}; having the code ARC refers to \texttt{setCode = 'ARC'}
};
\node at (0, 2) [draw, fill=pink!20, 
      rectangle, 
      rounded corners=10pt, 
      minimum width=16.5cm, 
      minimum height=5cm, 
      text width=15cm,
      align=left] 
{
Define the main table in the \texttt{FROM} clause: \texttt{`FROM sets AS T1`}.\\
Define the \texttt{JOIN} operation: \texttt{`INNER JOIN`}.\\
Define the table to join: \texttt{`set\_translations AS T2`}.\\
\textbf{\textcolor{red}{Define the join condition: \texttt{`ON T1.id = T2.id`}}}.\\
Define the main filtering condition in the \texttt{WHERE} clause: \texttt{`WHERE T1.mcmName = 'Archenemy'`}.\\
Add the additional filtering condition in the \texttt{WHERE} clause: \texttt{`AND T1.code = 'ARC'`}.\\
Select the column to be included in the final result: \texttt{`SELECT T2.language`}.\\[\baselineskip]

\textbf{[SQL]}\\
\texttt{SELECT T2.language FROM sets AS T1 INNER JOIN set\_translations AS T2 ON \textbf{\textcolor{red}{T1.id = T2.id}} WHERE T1.mcmName = 'Archenemy' AND T1.code = 'ARC'}
};
\node at (0, -3.5) [draw, fill=cyan!20, 
      rectangle, 
      rounded corners=10pt, 
      minimum width=16.5cm, 
      minimum height=5.5cm, 
      text width=15cm,
      align=left] 
{
Define the main table in the \texttt{FROM} clause: \texttt{`FROM sets AS T1`}.\\
Define the \texttt{JOIN} operation: \texttt{`INNER JOIN`}.\\
Define the table to join: \texttt{`set\_translations AS T2`}.\\
\textbf{\textcolor{green!65!black}{Define the join condition: \texttt{`ON T1.code = T2.setcode`.}}}\\
Define the main filtering condition in the \texttt{WHERE} clause: \texttt{`WHERE T1.mcmName = 'Archenemy'`}.\\
\st{Select the column to be included in the final result: \texttt{`SELECT T2.language`}.} \textbf{[BACK]}\\
Add the additional filtering condition in the \texttt{WHERE} clause: \texttt{`AND T2.setcode = 'ARC'`}.\\
Select the column to be included in the final result: \texttt{`SELECT T2.language`}.\\[\baselineskip]

\textbf{[SQL]}\\
\texttt{SELECT T2.language FROM sets AS T1 INNER JOIN set\_translations AS T2 ON \textbf{\textcolor{green!65!black}{T1.code = T2.setcode}} WHERE T1.mcmName = 'Archenemy' AND T2.setcode = 'ARC'}
};
\end{tikzpicture}
\caption{Comparison of error-free and \textit{RetrySQL} model generation. The external evidence is marked with yellow background, error-free response has a pink background, while the \textit{RetrySQL} response is denoted with blue background. The error-free model hallucinates the join condition \texttt{ON T1.id = T2.id}. The \textit{RetrySQL}-trained model generates the correct condition: \texttt{ON T1.code = T2.setcode}.}
\label{figure:generation_example_6}
\end{figure*}

\begin{figure*}[h]
\centering
\scalebox{0.9}{
\begin{tikzpicture}
\node[draw, fill=gray!10, 
          rectangle, 
          rounded corners=10pt, 
          minimum width=17cm, 
          minimum height=23cm, 
          text width=15cm,
          align=left, 
          drop shadow] 
{
};
\node at (0, 3) [draw, fill=pink!20, 
      rectangle, 
      rounded corners=10pt, 
      minimum width=16.5cm, 
      minimum height=16.5cm, 
      text width=15cm,
      align=left] 
{
Define the main table in the \texttt{FROM} clause: \texttt{`FROM superhero AS T1`}.\\
Define the first \texttt{JOIN} operation: \texttt{`INNER JOIN`}.\
Define the table to join: \texttt{`colour AS T2`}.\\
Define the join condition: \texttt{`ON T1.eye\_colour\_id = T2.id`}.\\
Define the second \texttt{JOIN} operation: \texttt{`INNER JOIN`}.\\
Define the table to join: \texttt{`colour AS T3`}.\\
Define the join condition: \texttt{`ON T1.hair\_colour\_id = T3.id`}.\\
Define the third \texttt{JOIN} operation: \texttt{`INNER JOIN`}.\\
Define the table to join: \texttt{`colour AS T4`}.\\
Define the join condition: \texttt{`ON T1.skin\_colour\_id = T4.id`}.\\
Define the fourth JOIN operation: \texttt{`INNER JOIN`}.\\
Define the table to join: \texttt{`gender AS T5`}.\\
Define the join condition: \texttt{`ON T1.gender\_id = T5.id`}.\\
Define the fifth \texttt{JOIN} operation: \texttt{`INNER JOIN`}.\\
Define the table to join: \texttt{`race AS T6`}.\\
Define the join condition: \texttt{`ON T1.race\_id = T6.id`}.\\
Define the sixth \texttt{JOIN} operation: \texttt{`INNER JOIN`}.\\
Define the table to join: \texttt{`publisher AS T7`}.\\
Define the join condition: \texttt{`ON T1.publisher\_id = T7.id`}.\\
Define the seventh \texttt{JOIN} operation: \texttt{`INNER JOIN`}.\\
Define the table to join: \texttt{`alignment AS T8`}.\\
Define the join condition: \texttt{`ON T1.alignment\_id = T8.id`}.\\
Define the main filtering condition in the \texttt{WHERE} clause: \texttt{`WHERE T2.colour = 'Blue'`}.\\
Add the additional filtering condition in the \texttt{WHERE} clause: \texttt{`AND T1.eye\_colour\_id = T2.id`}.\\
Add the additional filtering condition in the \texttt{WHERE} clause: \texttt{`AND T1.hair\_colour\_id = T3.id`}.\\
Add the additional filtering condition in the \texttt{WHERE} clause: \texttt{`AND T1.skin\_colour\_id = T4.id`}.\\
Add the additional filtering condition in the \texttt{WHERE} clause: \texttt{`AND T1.gender\_id = T5.id`}.\\
Add the additional filtering condition in the \texttt{WHERE} clause: \texttt{`AND T1.race\_id = T6.id`}.\\
Add the additional filtering condition in the \texttt{WHERE} clause: \texttt{`AND T1.publisher\_id = T7.id`}.\\
Add the additional filtering condition in the \texttt{WHERE} clause: \texttt{`AND T1.alignment\_id = T8.id`}.\\
Select the column to be included in the final result: \texttt{`SELECT COUNT(T1.id)`}.\\[\baselineskip]

\textbf{[SQL]}\\
\texttt{SELECT COUNT(T1.id) FROM superhero AS T1 INNER JOIN colour AS T2 ON T1.eye\_colour\_id = T2.id INNER JOIN colour AS T3 ON T1.hair\_colour\_id = T3.id INNER JOIN colour AS T4 ON T1.skin\_colour\_id = T4.id INNER JOIN gender AS T5 ON T1.gender\_id = T5.id INNER JOIN race AS T6 ON T1.race\_id = T6.id INNER JOIN publisher AS T7 ON T1.publisher\_id = T7.id INNER JOIN alignment AS T8 ON T1.alignment\_id = T8.id WHERE T2.colour = 'Blue' AND T1.eye\_colour\_id = T2.id AND T1.hair\_colour\_id = T3.id AND T1.skin\_colour\_id = T4.id AND T1.gender\_id = T5.id AND T1.race\_id = T6.id AND T1.publisher\_id = T7.id AND T1.alignment\_id = T8.id}
};
\node at (0, -8.37) [draw, fill=cyan!20, 
      rectangle, 
      rounded corners=10pt, 
      minimum width=16.5cm, 
      minimum height=5.75cm, 
      text width=15cm,
      align=left] 
{
Define the main table in the \texttt{FROM} clause: \texttt{`FROM superhero AS T1`}.\\
Define the \texttt{JOIN} operation: \texttt{`INNER JOIN`}.\\
Define the table to join: \texttt{`colour AS T2`}.\\
Define the join condition: \texttt{`ON T1.eye\_colour\_id = T2.id`}.\\
Define the main filtering condition in the \texttt{WHERE} clause: \texttt{`WHERE T2.colour = 'Blue'`}.\\
\st{Select the column to be included in the final result: \texttt{`SELECT COUNT(T1.id)`}.} \textbf{[BACK]}\\
Add the additional filtering condition in the \texttt{WHERE} clause: \texttt{`AND T1.eye\_colour\_id = T2.id`}.\\
Select the column to be included in the final result: \texttt{`SELECT COUNT(T1.id)`}.\\[\baselineskip]

\textbf{[SQL]}\\
\texttt{SELECT COUNT(T1.id) FROM superhero AS T1 INNER JOIN colour AS T2 ON T1.eye\_colour\_id = T2.id WHERE T2.colour = 'Blue' AND T1.eye\_colour\_id = T2.id}
};
\end{tikzpicture}
}
\caption{Comparison of error-free and \textit{RetrySQL} model generation. The error-free response has a pink background, while the \textit{RetrySQL} response is denoted with blue background. The error-free model generates many more reasoning steps than the \textit{RetrySQL}-trained model - it does not have the ability to backtrack, so it can only generate more steps to solve a given task.}
\label{fig:abs_reasoning_length_comparison_example}
\end{figure*}

\end{document}